\documentclass{article} 
\usepackage{iclr2023_conference,times}

\usepackage[pdftex]{graphicx}
\usepackage{hyperref}
\usepackage{url}
\usepackage{amsmath}
\usepackage{multirow}
\usepackage{color}
\usepackage{xcolor} 
\usepackage{colortbl}
\usepackage{wrapfig}
\newcommand{\etal}{\textit{et al}.~}
\newcommand{\ieno}{\textit{i}.\textit{e}.}
\newcommand{\egno}{\textit{e}.\textit{g}.} 
\usepackage{algorithm}
\usepackage{algorithmicx}
\usepackage{algpseudocode}
\newcommand{\tcb}{\textcolor{black}}

\newcommand{\tcr}{\textcolor{black}}

\newcommand{\ours}{\textit{CICF}}
\newcommand{\ourslong}{Confounder Identification-free Causal Visual Feature Learning}
\newcommand{\cluterthensample}{clustering-then-sampling}

\title{Confounder Identification-free Causal \\ Visual Feature Learning}

\author{Xin Li\textsuperscript{\rm 1}\footnotemark[1], Zhizheng Zhang\textsuperscript{\rm 2}\footnotemark[2], Guoqiang Wei\textsuperscript{\rm 1}\footnotemark[1],  Cuiling Lan\textsuperscript{\rm 2}\footnotemark[2], Wenjun Zeng\textsuperscript{\rm 3}, Xin Jin\textsuperscript{\rm 3}, Zhibo Chen\textsuperscript{\rm 1}\footnotemark[2] \\
\textsuperscript{\rm 1}University of Science and Technology of China \\
\textsuperscript{\rm 2}Microsoft Research Asia \\ \textsuperscript{\rm 3}EIT Institute for Advanced Study \\
\texttt{\{lixin666, wgq7441, jinxustc\}@mail.ustc.edu.cn} \\
\texttt{\{zhizzhang, culan\}@microsoft.com} \\
\texttt{zengw2011@hotmail.com}, \texttt{chenzhibo@ustc.edu.cn}
}

\iclrfinalcopy 
\begin{document}
\definecolor{mygray}{gray}{.9}

\maketitle
\renewcommand{\thefootnote}{\fnsymbol{footnote}} 
\footnotetext[1]{This work was done when Xin Li and Guoqiang were interns at MSRA.} 
\footnotetext[2]{Corresponding author.}
\begin{abstract}
Confounders in deep learning are in general detrimental to model's generalization where they infiltrate feature representations. Therefore, learning causal features that are free of interference from confounders is important. Most previous causal learning-based approaches employ back-door criterion to mitigate the adverse effect of certain specific confounder\tcb{s}, which require the explicit identification of confounder\tcb{s}. However, in real scenarios, confounders are typically diverse and difficult to be identified. 
In this paper, we propose a novel \textbf{C}onfounder \textbf{I}dentification-free \textbf{C}ausal Visual \textbf{F}eature Learning (\ours) method, which obviates the need for identifying confounders.
\ours~models the interventions among different samples based on the front-door criterion, and then approximates the global-scope intervening effect based on the instance-level intervention from the perspective of optimization. 
In this way, we aim to find a reliable 
optimization direction, which \tcb{eliminates the confounding effects} of confounders, to learn causal features. 
Furthermore, we uncover the relation between \ours~and the popular meta-learning strategy MAML \citep{finn2017modelMAML}, and provide an interpretation of why MAML works from the theoretical perspective of causal learning for the first time. Thanks to the effective learning of causal features, our \ours~enables models to have superior generalization capability. Extensive experiments on domain generalization benchmark datasets demonstrate the effectiveness of our \ours, which achieves the state-of-the-art performance. 
\end{abstract}

\section{Introduction}
\label{sec:intro}

Deep learning excels at capturing correlations between the inputs and labels in a data-driven manner, which has achieved remarkable successes on various tasks, such as image classification, object detection, and question answering \citep{liu2021swin,he2016deepResNet,redmon2016youYOLO,he2017maskRCNN,antol2015VQA}. Even so, in the field of statistics, \textit{correlation is in fact not equivalent to causation}~\citep{pearl2016causal}.  For example, when tree branches usually appear together with birds in the training data, deep neural networks (DNNs) are easy to mistake features of tree branches as the features of birds. A close association between two variables does not imply that one of them causes the other. Capturing/modeling correlations instead of causation is at high risk of allowing various confounders to infiltrate into the learned feature representations. 
When affected by intervening effects of confounders, a network may still make correct predictions when the testing and training data follow the same distribution, but fails when the testing data is out of distribution. This harms the generalization capability of learned feature representations.
Thus, learning causal feature, where the interference of confounders is excluded, is important for achieving reliable results.

\begin{figure}[t]
    \begin{center}
\includegraphics[width=0.68\linewidth]{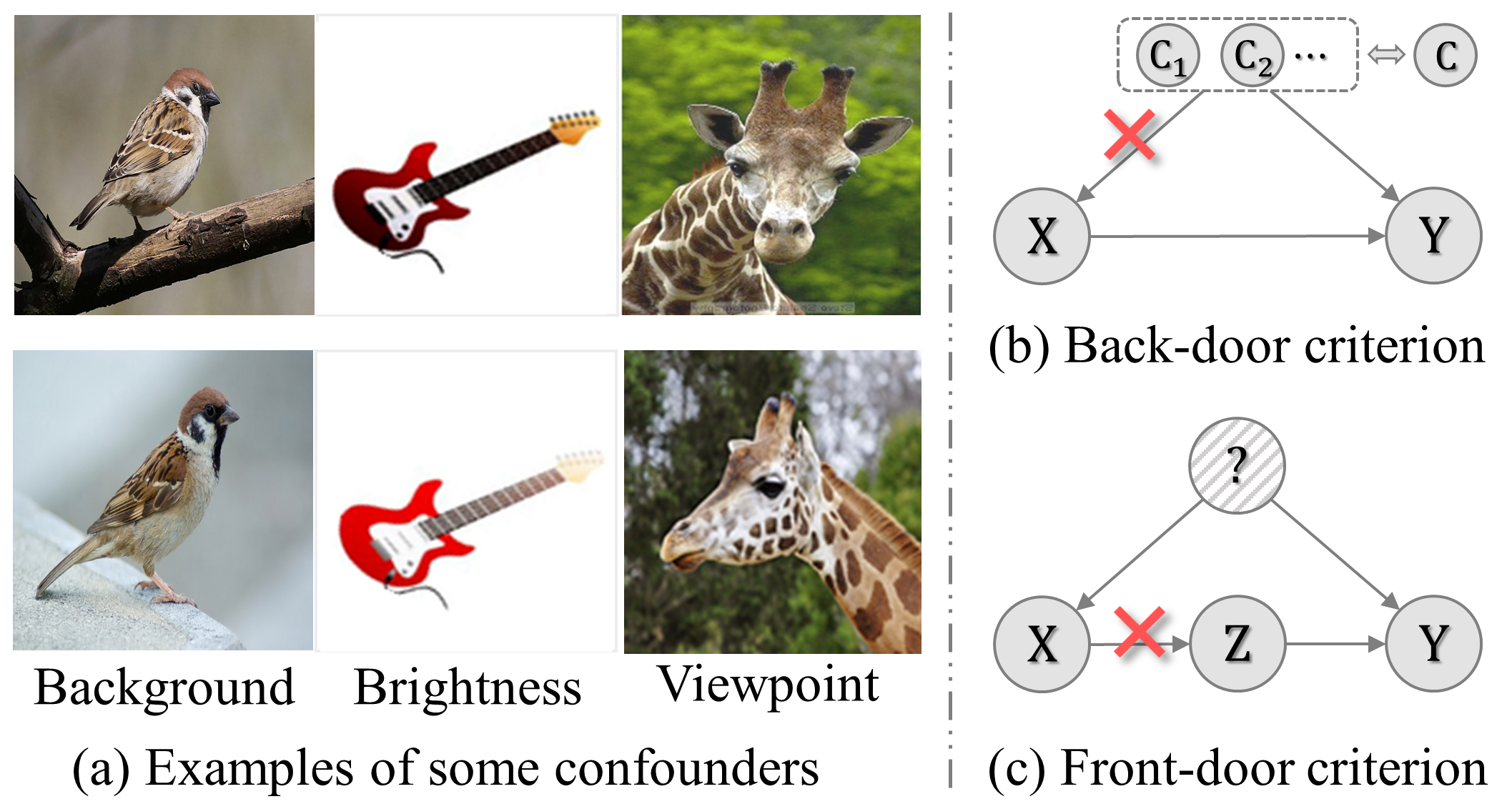}
    \end{center}    
    \caption{(a) Examples of some confounders, which may lead to learning biased features.
    (b) Back-door criterion in causal inference, where the counfunders are accessible.
    (c) Front-door criterion in causal inference, where the confounders are inaccessible.}
    \label{fig:causal}
    \vspace{-6mm}
\end{figure}

As shown in Fig.~\ref{fig:causal}, confounders $C$ bring a spurious (non-causal) connection $X\xleftarrow{}C\xrightarrow{}Y$ between samples $X$ and their corresponding labels $Y$. A classical example to shed light on this is that we can instantiate $X, Y, C$ as the sales volume of ice cream, violent crime and hot weather. Seemingly, an increase in ice cream sales $X$ is correlated with an increase in violent crime $Y$. 
However, the hot weather is the common cause of them, which makes an increase in ice cream sales to be a misleading factor of analyzing violent crime.
Analogically, in deep learning, once the misleading features/confounders are captured, the introduced biases may be mistakenly fitted by neural networks, thus leading to the detriment of the generalization capability of learned features.  
In theory, we expect DNNs to model the causation between $X$ and $Y$.
Deviating from such expectation, the interventions of confounders $C$ make the learned model
implicitly condition on $C$. 
This makes that the regular feature learning does not approach the causal feature learning.
To learn causal features, previous studies~\citep{yue2020interventional,zhang2020causal,wang2020visual} adopt the backdoor criterion \citep{pearl2016causal} to explicitly identify confounders that should be adjusted for modeling intervening effects. However, they can only exploit the confounders that are accessible and can be estimated, leaving others still intervening the causation learning. Moreover, in many scenarios, confounders are unidentifiable or their distributions are hard to model \citep{pearl2016causal}.

Theoretically, front-door criterion\citep{pearl2016causal} does not require identifying/explicitly modeling confounders. It introduces an intermediate variable $Z$ and transfers the requirement of modeling the intervening effects of confounders $C$ on $X \to Y$ to modeling the intervening effects of $X$ on $Z \to Y$.
Without requiring explicitly modeling confounders, the front-door criterion is inherently suitable for wider scenarios. 
However, how to exploit the front-door criterion for causal visual feature learning is still under-explored.

In this paper, we design a Confounder Identification-free Causal visual Feature learning method (\ours). Particularly, \ours~models the interventions among different samples based on the front-door criterion, and then approximates the global-scope intervening effect based on the instance-level interventions from the perspective of optimization. 
In this way, we aim to find a reliable optimization direction, 
which \tcr{eliminates the confounding effects} of confounders, to learn causal features. 
There are two challenges we will address for \ours.
1) How to model the intervening effects from other samples
on a given sample in the training process.
2) How to estimate the global-scope intervening effect across all samples in the training set to find a  suitable optimization direction.

As we know, during training, each sample intervenes others through its effects on network parameters by means of gradient updating. 
Inspired by this, we propose a gradient-based method to model the intervening effects on a sample from all samples to learn causal visual features. 
However, it is intractable to involve such modeled global-scope intervening effects in the network optimization, which requires a traversal over the entire training set and is costly. 
To address this, we propose an efficient cluster-then-sample algorithm to approximate the global-scope intervening effects for feasible optimization. 
Moreover, we revisit the popular meta-learning method Model-Agnostic Meta-Learning (MAML) \citep{finn2017modelMAML}. We surprisingly found that our \ours~can provide an interpretation on why MAML works well from the perspective of causal learning: MAML tends to learn causal features.
We validate the effectiveness of our \ours~on the Domain Generalization (DG) \citep{wang2021generalizingDGSurvey,zhou2021domainDGSurveyzhou} task and conduct extensive experiments on the PACS, Digits-DG, Office-Home, and VLCS datasets. Our method achieves the state-of-the-art performance.

\section{Related Work}
\noindent\textbf{Causal Inference} aims at pursuing the causal effect of a particular phenomenon by removing the interventions from the confounders \citep{pearl2016causal}.  
Despite its success in economics~\citep{rubin1986statistics}, statistics~\citep{rubin1986statistics,imbens2015causal} and social science~\citep{murnane2010methods}, big challenges present when it meets machine learning, \ieno, how to model the intervention from the confounders and how to establish the causal model. 
A growing number of works have moved a step forward by taking advantage of the back-door criterion \citep{pearl2016causal} on various tasks, 
\egno, few-shot classification~\citep{yue2020interventional}, vision-language task~\citep{wang2020visual}, domain adaptation~\citep{yue2021transporting}, class-incremental learning~\citep{hu2021distilling}, and semantic segmentation~\citep{zhang2020causal}. 
Limited by the back-door criterion, most of them are required to identify and model the distributions of the confounders. However, this may be challenging in the real world because confounders are typically diverse and usually appear implicitly.
To get rid of the dependency on confounders,~\citet{yang2021causal}, for the first time, propose to utilize the front-door criterion to establish a causal attention module for vision-language task. However, it still requires the modeling of intervention in the testing stage, which is complicated.

In contrast, this work is the first attempt to 
apply the front-door criterion for learning causal visual features by considering the intervention among samples. Ours improves the generalization ability of DNNs 
from the optimization perspective and is confounder identification-free. 

\noindent\textbf{Model Generalization} plays a prominent role for DNNs to be applied in real-world scenarios.
To improve the performance on the testing dataset which has distribution shift \citep{sun2016returnCORAL} with training data, 
various domain generalization (DG) \citep{muandet2013domainFIRSTDG,zhou2021domainDGSurveyzhou,wang2021generalizingDGSurvey,shen2021towardsSurveyCUIPENG,wei2021metaalign} methods have been proposed. 
In general, these methods can be divided into three categories, \ieno, domain-invariant representation learning, data or feature manipulation, and meta-learning.
The first category intends to learn domain-invariant features that follow the same distributions \citep{muandet2013domainFIRSTDG,li2018domainDGMMDAAE,taori2020measuring,li2018deepDeepConditionIDN,motiian2017unifiedCCSA,mahajan2021domainDGCausalMatch,jin2020featureFAR}. The second category aims to improve the generalization ability of models through enriching the diversity of source domains, either in image space (\egno, CrossGrad \citep{shankar2018generalizingCrossGrad}, DDAIG \citep{zhou2020deep} and M-ADA \citep{qiao2020learningMADA}), or feature space (MixStyle~\citep{zhou2021domainMixstyle} and RSC \citep{huang2020selfRSC}). 
Another new line of DGs utilize meta-learning as training strategy~\citep{zhao2021learningMetaReid,liu2020shapeShapeMeta,li2018learningMLDG,li2020sequentialSMLDG,wei2021metaalign,balaji2018metareg}. MAML~\citep{finn2017modelMAML} takes the advantage of meta-learning to find a good parameters initialization for fast adaptation to new tasks. Following MAML, \citet{li2018learningMLDG,dou2019domainMASF,li2020sequentialSMLDG,balaji2018metareg} introduce meta-learning into DG to simulate domain shift or learn domain-invariant parameters regularizer during training. 
Other variants of DGs exploit episodic training~\citep{li2019episodic} and ensemble learning~\citep{zhou2021domainDGensemble,seo2020learningDomainSpecificNorm, cha2021domain}. 

In this paper, from a new perspective, we propose a scheme for model generalization termed as 
\ourslong~(\ours).

\section{Proposed Method}
In this section, we first depict a supervised learning process in  a causal graph \citep{pearl2009causality}, and uncover the stumbling effects of confounders which prevent the achievement of high generalization capability of models in Sec.~\ref{sec:theory}. Then, in Sec.~\ref{sec:cicf}, based on the front-door criterion, we elaborate our \ourslong (\ours) from two perspectives, respectively as how to model mutual intervening effects between different instances and how to approximate such intervening effects from the global scope. Furthermore, we describe our \ours~which alleviates the intervening effects from the optimization perspective in Sec.~\ref{sec:cicf}.
In Sec.~\ref{sec: discussion}, we uncover the relation between our \ours~and the popular meta-learning strategy MAML~\citep{finn2017modelMAML}, and provide an interpretation of why MAML works from the theoretical perspective of causal inference.

\subsection{Problem Definition and Analysis}
\label{sec:theory}

\begin{figure}[t]
    \centering
    \includegraphics[width=0.96\linewidth]{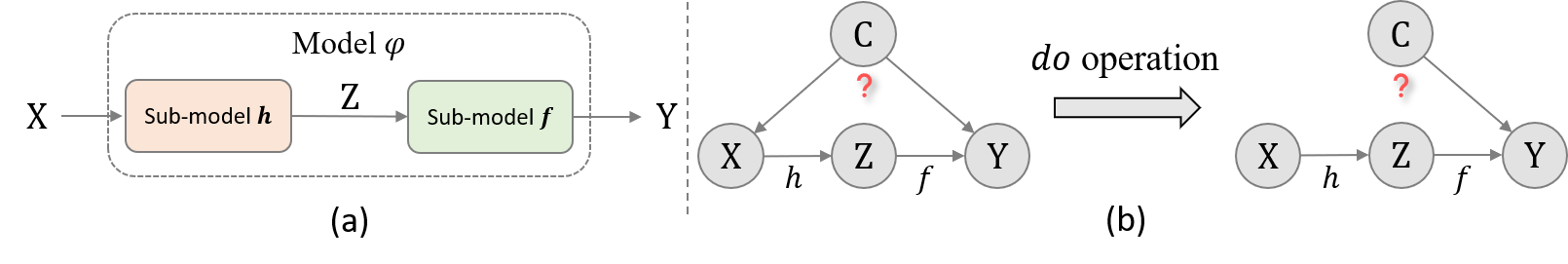}
    \caption{Causal graph in supervised learning. (a) A network/model $\varphi$ consists of sub-models $h$ and $f$. $X$ and $Y$ denote the input and label, respectively. (b) Causal graph where \emph{do} operation aims to remove the intervening from confounder $C$.}
    \label{fig:SCM_for_Training}
    \vspace{-3mm}
\end{figure}

Given a training dataset with input and label pairs $\{X, Y\}$, the goal of training Deep Neural Networks is to learn/capture the causation between input samples $X$ and labels $Y$, \ieno, 
the conditional probability $P(Y|do(X))$. As shown in Fig.~\ref{fig:SCM_for_Training} (a), we parameterize the network as $\varphi$ and separate it into two successive parts, \ieno, $h$ and $f$. 

DNNs capture label-associated features which are not necessarily the casual ones due to the intervening effects of confounders, such as background, brightness, and viewpoint. We denote the intermediate features and confounders as $Z$ and $C$, respectively.
Fig. \ref{fig:SCM_for_Training} illustrates the relations in a causal graph. Intervened by the confounders, the conditional probability $P(Y|X)$ learned by the model $\varphi$ actually involves two paths, \ieno, $X\xrightarrow{}Z\xrightarrow{}Y$ and $X\xleftarrow{}C\xrightarrow{}Y$. 
$X\xrightarrow{}Z\xrightarrow{}Y$ denotes the expected
causal effect from the input samples $X$ to their corresponding labels $Y$. 
The path $X\xleftarrow{}C\xrightarrow{}Y$ denotes the non-causal correlation between $X$ and $Y$ due to their common cause $C$,
which may introduce biases into the learning of $P(Y|X)$ and thus affect the generalization capability of model $\varphi$. 

Intuitively, it is crucial to get rid of the harmful bias from those confounders for causal feature learning. In the literature of causal inference \citep{pearl2016causal}, the confounding effects from confounders can be removed through the \textit{do} operation \citep{pearl2016causal} by cutting off the connection from $C$ to $X$, as illustrated in Fig. \ref{fig:causal} (b). With the definition of \textit{do} operation, the real causation from $X$ to $Y$ can be formulated by $P(Y|do(X))$. The objective of our \ours~is to learn features representation conforming to $P(Y|do(X))$.

\noindent\textbf{Back-door criterion.}
In previous works \citep{yue2020interventional,wang2020visual,hu2021distilling}, when $C$ is identifiable, the back-door criterion \citep{pearl2016causal} is typically utilized to achieve the \textit{do} operation as:
\begin{equation}
\centering
 P(Y|do(X=x))=\sum_{c}P(Y|X=x, C=c)P(C=c),
 \label{equ:back-door}
\end{equation}
which acquires access to the distributions of all confounders ${c\in C}$. However, in many scenarios, the intervening effects are caused by unobservable or implicit factors. It is not feasible to identify the distribution of $C$ during training, limiting the usage of the back-door criterion.

\noindent\textbf{Front-door criterion.}
For unidentifiable confounders, the \textbf{F}ront-door criterion \citep{pearl2016causal} provides us with a more practical alternative to \textbf{E}stimate the \textbf{I}ntervening \textbf{E}ffect, called \textit{FEIE}, eschewing the identification of confounders $C$. Specifically, it introduces an intermediate variable $Z$ to help assess the effect of $X$ on $Y$, \ieno, $P(Y|do(X))$, which can be formulated as:
\begin{align}
    \centering
    P(Y|do(X=x))=\sum_{z}P(Z=z|X=x)\sum_{\Tilde{x}\in X}^{} P(Y|Z=z, \Tilde{x}) P(\Tilde{x}),
    \label{equ:front-door}
\end{align}
where $Z=h(X)$, $\Tilde{x}\in X$ denotes a sample from training data. Note that the effect of $X$ on $Z$ is identifiable because they have no common causes. In other words, there is no backdoor path from $X$ to $Z$. Thus, we have $P(Z=z|do(X=x))=P(Z=z|X=x)$.

Front-door criterion is attractive for eliminating interventions. However, it is still under-explored for visual feature learning, where there is a lack of a simple and practical mechanism to exploit this theory for enhancing the generalization capability of models.

\subsection{Confounder Identification-free Causal Visual Feature Learning}
\label{sec:cicf}
In this paper, we aim to achieve \textit{Confounder Identification-free Causal Visual Feature Learning}, obviating the need for confounder identification.
As indicated by Eq. (\ref{equ:front-door}), thanks to 
the front-door criterion, we do not need to identify and explicitly model confounders $C$. Despite this, it still imposes a challenge on how to accurately model the term $\sum_{\Tilde{x}}P(Y|Z=z, \Tilde{x})$ in the network training process. We treat the first part of $\varphi$ as the model $h$ to obtain the intermediate variable $z$, \ieno, $z=h(x)$. Because the parameters of $h$ are fixed in the inference stage and $z=h(x)$ is known given any $x\in X$, $P(Z=z|X=x)$ is equal to $1$ \footnote{We provide proof in Appendix~\ref{sec:proof} that this satisfies the front-door criterion.}.  Thus, we can re-write Eq.~(\ref{equ:front-door}) as:
\begin{equation}
 \centering
P(Y|do(X=x))=\sum_{\Tilde{x}\in X}^{}P(Y|Z=h(x),\Tilde{x})P(\Tilde{x}), 
\label{equ:front-door_stage2}  
\end{equation}
where $P(Y|Z=h(x), \Tilde{x})$ denotes the mutual intervening effects to the causation path $Z\xrightarrow{}Y$ from another sample $\Tilde{x}$. 
With the summation operation, Eq. \ref{equ:front-door_stage2} represents the \textit{global} intervening effects accumulated from all samples in the training data. 
We will describe the instantiations of $P(Y|Z=h(x), {\tilde{x}})$ and the accumulation in Eq.~(\ref{equ:front-door_stage2}) respectively as below.

\noindent\textbf{A Gradient-based Instantiation of \textit{FEIE}.}
\label{sec:optim_intervention}
Referring to the practices in prior works \citep{yue2020interventional,wang2020commonsense} upon the back-door criterion, a straightforward method for modelling $P(Y|Z=h(x), \Tilde{x})$ is to directly concatenate $z=h(x)$ and the feature of $\Tilde{x}\in X$ before feeding them into $f$. However, this method would easily lead to a trivial solution once the information of $\Tilde{x}$ is ignored by the layers of the neural networks. In contrast, we propose to explicitly model the intervening effects of $\Tilde{x}$ on $Z\xrightarrow{}Y$ with a gradient-based instantiation. We notice that, in the training process, \textit{the influence of one instance on others can be reflected on the 
parameters updating with the gradient obtained based on this instance}. 
Therefore, for a given sample $x$, we propose to explicitly model the intervening effects $P(Y|Z=h(x),\tilde{x})$ of another sample $\tilde{x}$ on $x$ through $f$ as:
\begin{equation}    
    P(Y|Z=h(x), \tilde{x}) =f_{\theta_{\tilde{x}}} (Z=h(x)), 
    ~where ~ \theta_{\tilde{x}}=\theta - \alpha g_{\tilde{x}}, 
     ~{g_{\tilde{x}}}={\nabla}_{\theta}{\mathcal{L}}(f_{\theta}(h({\tilde{x}}),{\tilde{y}})),
    \label{equ:gradient_intervention}
\end{equation}
$f_{\theta}$ and $f_{\theta_{\tilde{x}}}$  
denote the model $f$ before and after the parameters updating respectively,
$g_{\tilde{x}}$ denotes the calculated gradient 
with respect to the sample $\tilde{x}$ and its label $\tilde{y}$. $\mathcal{L}$ and $\alpha$ represent the loss of cross entropy and learning rate, respectively. Incorporating Eq. (\ref{equ:gradient_intervention}) into Eq. (\ref{equ:front-door_stage2}), we have Eq.~(\ref{equ:front-door_gradient}) as below to explicitly eliminate the interventions of all samples on the sample $x$ as:
\begin{equation}
    \centering
    P(Y|do(X=x))=\sum_{\Tilde{x}\in X}^{}f_{\theta_{\Tilde{x}} }(Z=h(x))P({\Tilde{x}}).
    \label{equ:front-door_gradient}
\end{equation}
\noindent\textbf{Global-scope Intervening Effects Approximation.} 
\label{sec:global-scope-intervention}
With the above introduced gradient-based instantiation, the globally accumulative intervening effects from {all the training samples} can be estimated by a traversal on $X$, which, however, is time- and memory-consuming in practice. To achieve an efficient estimation in the global scope, we apply the first-order Taylor's expansion on Eq. (\ref{equ:front-door_gradient}):
\begin{align}
     P(Y|do(X=x)) 
     &=\!\!\sum_{\Tilde{x}\in X}^{} 
     \!\!\left[ f_\theta (h(x))\!-\!{\alpha}g_{\Tilde{x}} \nabla_\theta f_{\theta}(h(x))\!+\! 
o\left( \nabla_\theta f_{\theta}(h(x)) \right) \right]P(\Tilde{x}) 
\notag\\&\approx  f_{\theta}(h(x))-\alpha(\sum_{{\Tilde{x}}\in X}^{}g_{\Tilde{x}}P({\Tilde{x}})) \nabla_\theta f_{\theta}(h(x)).
\label{equ:taylor}
\end{align}
Eq.~(\ref{equ:taylor}) reveals that the key to estimating $P(Y|do(X))$ lies in computing the global-scope 
gradient $g_\dagger=\sum_{{\Tilde{x}}\in X}g_{\Tilde{x}}P({\Tilde{x}})$ 
over all $\Tilde{x}\in X$ accumulated via weighted sum \tcr{with $P({\Tilde{x}})$ as the weight}.

However, it is intractable to directly  compute the global-scope gradient $g_\dagger$ by traversing over all the training samples. \emph{Alternatively, we can traverse over a sampled small subset that shares the similar data distribution to that of all the training data.} As we know, when the training data are unbalanced and diverse, random sampling of a small subset would result in bias that mismatches the distribution of the dataset, leading to an inaccurate estimation of the global-scope gradient.
To \tcr{better} estimate the data distribution and thus \tcr{approach} the global-scope gradient, we propose a sampling strategy dubbed as \textit{clustering-then-sampling}. 
More discussion/analysis can be found in the \textbf{Appendix}~\ref{sec:more_details}.
Concretely, we first cluster the training samples of each class in the dataset into $K$ clusters with $K$-means algorithms~\citep{pelleg2000xKmeans} and totally obtain $K^\dagger$ clusters for the whole training data.

It is noteworthy that 
we found the samples in each cluster usually have similar gradient directions in optimization.  
Thus we represent each cluster with fewer samples randomly sampled from the same cluster, avoiding traversing over all the data. 
Then, the global-scope gradient $g_\dagger$ can be approximated with weighted sum over the sampled $M=$ $\sum_{k=1}^{K^\dagger}\!\!N_k$ samples from $K^\dagger$ clusters: 
\begin{equation}
    \centering
    g_\dagger=\frac{1}{M}\sum_{k=1}^{K^\dagger}\sum_{j=1}^{N_{k}} g_{\Tilde{x}_{j,k}\in K_k},
    \label{equ:cluster_our}
\end{equation}
where 
$N_k$ is the number of instances sampled from the $k$-th cluster (being proportional to the size of this cluster), $g_{\Tilde{x}_{j,k}\in K_k}$ denotes the gradients of the sample ${\Tilde{x}_{j,k}\in K_k}$. 
Combined with Eq. (\ref{equ:cluster_our}), 
we rewrite Eq.~(\ref{equ:taylor}) as: 
\begin{align}
    P(Y&|do(X=x)) 
     \approx f_{\theta}(h(x))-\alpha g_\dagger\nabla_\theta f_{\theta}(h(x)).
     \label{equ: equ10}
\end{align}
\noindent\textbf{Causal Visual Feature Learning.}
\label{sec: summary}
\emph{Based on the above theoretical analysis and the proposed intervention \tcr{approximation} strategy, the intractable causal conditional probability  $P(Y|do(X))$ can be approximated \tcr{based on Eq.(\ref{equ: equ10})}, without requiring the identification of confounders $C$.}
Actually, the Eq. (\ref{equ: equ10}) can be viewed as the first-order Taylor's expansion of \tcr{$f_{\theta- \alpha {g_\dagger}} (h(x))$}. Thus, we have:
\begin{equation}
\centering
    P(Y|do(X=x)) \approx
    f_{\theta- \alpha {g_\dagger}}(h(x)), 
    \label{equ:two_stage}
\end{equation}
here let $\theta_{\dagger} = \theta- \alpha {g_\dagger}$ (the parameters of $f$), which are updated with the global-scope gradient $g_\dagger$.
The output of a model is thus denoted as $\hat{y}_{do(x)}=f_{\theta_{\dagger}} (h(x))$, 
which has been aware of the global-scope interventions from all other samples 
on the current sample $x$ based on such global-scope gradient updated model (\emph{i.e.}, $f_{\theta_\dagger}$). 
Then, we can train an \textit{unbiased} model $f$ to learn the causal visual features with the loss of cross-entropy: 
\begin{equation}
    \centering
    \mathcal{L}_{\ours} = 
    \sum_{x\in X}\mathcal{L}_{ce}\left(f_{\theta_\dagger}(h(x)), y \right),
    \label{equ:celoss}
\end{equation}
where $y$ is the corresponding ground-truth label for $x$.
The overall algorithm of \textit{\ourslong} is described in Alg.~\ref{alg:cicf} of Appendix.

Note that clustering-then-sampling is better than random sampling to \tcr{approach} the distribution of the training dataset, thereby being capable of approximating the global-scope gradient more accurately.
We have theoretically analyzed that  \textit{\cluterthensample} has a more minor standard error (SE) for estimating the distribution of all the training data than  random sampling, \ieno, $SE_{ours} < SE_{random}$ in the \textbf{Appendix}~\ref{sec:more_details}. 
This demonstrates that our \textit{\cluterthensample} is a more efficient and more accurate strategy to estimate the data distribution and then the global-scope gradient. 

\vspace{-2mm}
\subsection{Discussion}
\label{sec: discussion}

In this section, we will provide an analysis and comparison between
our \ours~and MAML \citep{finn2017modelMAML}. For the first time, we interpret why MAML works from a causal learning perspective, which is supported by our analysis in the previous subsections.

In the seminal work MAML~\citep{finn2017modelMAML}, Finn \etal propose a model-agnostic meta-learning strategy that treats a batch of data as meta-train and another batch of data as meta-test for optimization.
Particularly, given $T$ sets of data $\{D_{tr}^t,D_{te}^t\}_{t=1}^{T}$ corresponding to $T$ tasks $\{\mathcal{T}_t\}_{t=1}^T$, where $D_{tr}^{t}$ and $D_{te}^t$ denote meta-train and meta-test data respectively, 
the loss function of MAML for optimization can be represented as:
\begin{equation}
   \centering
\mathcal{L}_{MAML}=\sum_t\mathcal{L}\left(f_{\theta_{tr}^t}(X_{te}^t), Y_{te}^t\right),
    \label{eq: maml}
\end{equation}
where $\theta_{tr}^t$ refers to the parameter 
virtually updated with the gradient $g_{tr}^t$ calculated on $D_{tr}^t$, \ieno, 
$\theta_{tr}^t\leftarrow\theta - \alpha {g}_{tr}^t$, 
which is treated as meta-train task.
The optimization on ${D_{te}^t}$ is treated as meta-test task, where the parameters are updated as $\theta_{te}^t\leftarrow\theta - \alpha \nabla_{\theta} \mathcal{L}_{MAML}$.
They interpret why MAML works from the perspective that it can provide a good parameter initialization which is robust for fast adaptation to new data. 
However, there is a lack of theoretical analysis and support in \citet{finn2017modelMAML}.
Based on our analysis in Section \ref{sec:cicf}, for the first time, we have \textit{a new understanding of 
the previous uses of MAML ~\citep{finn2017modelMAML,li2018learningMLDG} (see Eq. (\ref{eq: maml})) from the perspective of causal inference}.
$\mathcal{L}\left(f_{\theta_{tr}^t}(X_{te}^t), Y_{te}^t\right)$ in Eq.~(\ref{eq: maml}) actually models the intervention from meta-train data $D_{tr}^t$ to meta-test data $D_{te}^t$ within the task $t$ and endeavors to eliminate such local data modeled intervention. However, as revealed by our theoretical analysis in Section \ref{sec:cicf}, learning reliable causal features requires the capturing and modeling of interventions from all the samples (\ieno, global interventions). There is no such solution in the previous works while we provide a practical and 
efficient one to model and eliminate the global-scope interventions in this paper. This enables reliable causal feature learning and promotes the achievement of higher generalization capability of models.
\begin{figure*}[t]
    \centering
    \includegraphics[width=0.95\linewidth]{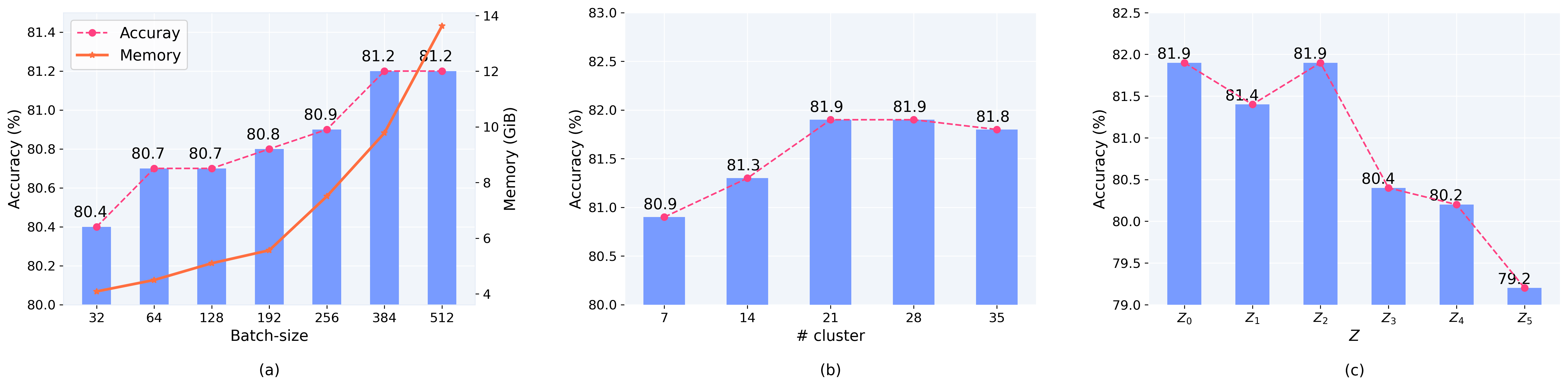}
    \caption{Ablation study on the influence from (a) batch-size, (b) the number of clusters, for the global-scope intervention estimation, and (c) different choices of $Z$ (c.f. Fig. \ref{fig: ablation_layer}). \tcr{$\{Z_0, ..., Z_5\}$ denotes the features from the shallow layer to the deep layer.}
    All results are averaged over four domains on PACS.}
    \vspace{-3mm}
    \label{fig:ablation}
\end{figure*}

\begin{table}[htp]
\centering
\caption{Classification accuracy ($\%$) of different DG methods on PACS with ResNet-18 and Digits-DG with a CNN backbone ~\citep{zhou2021domainMixstyle}. $\ddagger$: our reimplemented results using the official code are different from the ones reported in the original paper.}
\setlength{\tabcolsep}{0.8mm}{
\begin{tabular}{c|cccc|c|c|ccccc}
\hline \hline
\multirow{2}{*}{Method} & \multicolumn{5}{c|}{PACS}              &  & \multicolumn{5}{c}{Digits-DG}                                 \\ \cline{2-6} \cline{8-12} 
                        & A  & C & P & S & Avg. &  & MINIST & MINIST-M & SVHN & \multicolumn{1}{c|}{SYN} & Avg. \\ \cline{1-6} \cline{8-12} 
                        
MMD-AAE                 & 75.2 & 72.7    & 96.0  & 64.2   & 77.0 &  &  96.5      &   58.4       &  65.0    & \multicolumn{1}{c|}{78.4}    &   74.6   \\
CCSA                   & 80.5 & 76.9    & 93.6  & 66.8   & 79.4 &  &  95.2      &   58.2       &  65.5    &  \multicolumn{1}{c|}{79.1}    &   74.5   \\
JiGen                  & 79.4 & 75.3    & 96.0  & 71.6   & 80.5 &  &  96.5      &   61.4       &  63.7    & \multicolumn{1}{c|}{74.0}    &  73.9    \\
CrossGrad              & 79.8 & 76.8    & 96.0  & 70.2   & 80.7 &  &  96.7      &   61.1       &  65.3    & \multicolumn{1}{c|}{80.2}    &  75.8    \\
MLDG               & 79.5 & 77.3    & 94.3  & 71.5   & 80.7 &  &    94.7    & 60.3        & 61.5    & \multicolumn{1}{l|}{75.4}    & 72.6     \\
MASF    & 80.3 & 77.2  & 95.0  & 71.7 & 81.1 & & - & - & - &\multicolumn{1}{c|}{-}& - \\
MetaReg                & 83.7 & 77.2    & 95.5  & 70.3   & 81.7 &  &  -      &     -     & -     & \multicolumn{1}{c|}{-}    &     - \\
RSC               & 83.4 & 80.3    & 96.0  & 80.9   & 85.2 &  & -  & -   & -    & \multicolumn{1}{c|}{-}   & -     \\ 
MatchDG                & 81.3 & \textbf{80.7}    & \textbf{96.5}  & 79.7   & 84.6 &  & -       & -        & -     &  \multicolumn{1}{c|}{-}    & -    \\
MixStyle$^\ddagger$               & 83.0 & 78.6    & 96.3  & 71.2   & 82.3 &  &  96.5      &    63.5      &   64.7   & \multicolumn{1}{c|}{81.2}    &   76.5   \\   
FACT    & \textbf{85.4} & 78.4 & 95.2 & 79.2 & 84.5 & & \textbf{97.9} & 65.6 & 72.4 & \multicolumn{1}{c|}{\textbf{90.3}} & 81.5 \\

\hline \hline
ERM & 77.0 & 75.9    & 96.0  & 69.2   & 79.5 &  &    95.8    &    58.8      &  61.7    & \multicolumn{1}{c|}{78.6}    &  73.7    \\
ERM+MAML  & 77.0 &  74.5   & 94.8 & 72.1  & 79.6 &  &     96.0  &  63.1       & 65.0   & \multicolumn{1}{c|}{81.1}    &    76.5 \\
\rowcolor[gray]{0.9}
ERM+\ours          & 80.7 & 76.9    & 95.6  & 74.5   & 81.9 &  &    95.8    &    63.7      &   65.8   & \multicolumn{1}{c|}{80.7}    &  76.5    \\\cline{1-6} \cline{8-12}
ERM$^*$            & 82.5 & 74.2    & 95.4  & 76.5   & 82.1 &  &  96.1      &  65.0         &   73.0    & \multicolumn{1}{c|}{84.6}    &  79.7     \\
ERM$^*$+MAML            & 81.8 & 73.2    & 94.8  & 75.7   & 81.4 &  & 96.2       &67.0          & 74.0     & \multicolumn{1}{c|}{84.1}    & 80.3     \\
\rowcolor[gray]{0.9}
ERM$^*$+\ours       & 84.2 & 78.8    & 95.1  & \textbf{83.2}   & \textbf{85.3} &  &   95.6     &   \textbf{68.8}       &  \textbf{76.5}    & \multicolumn{1}{c|}{86.0}    &  \textbf{81.7}   \\ \cline{1-6} \cline{8-12}
 \hline \hline
\end{tabular}}
\label{tab:pacs}
\vspace{-3mm}
\end{table}
\vspace{-3mm}
\section{Experiments}
To validate the effectiveness of our \ours, we apply it on the Domain Generalization (DG) \citep{zhou2021domainDGSurveyzhou,wang2021generalizingDGSurvey} task, where the models are expected to learn the unbiased causal features to be generalized to different domains. 
We describe the datasets and implementation details in Sec. \ref{sec: datasets}. Then we clarify the effectiveness of each component of our \ours~in Sec. \ref{sec: ablation}, and compare with previous methods
in Sec.  \ref{sec: comparisons}. Finally, we qualitatively show that \ours~captures the causal features in Sec. \ref{sec: feat}.
\subsection{Datasets and Implementation Details}
\label{sec: datasets}
\textbf{Datasets.} We evaluate our method on four commonly used benchmark datasets (\ieno, PACS~\citep{li2017deeperPACS}, Digits-DG, Office-Home~\citep{venkateswara2017deepOffice-Home} and VLCS~\citep{torralba2011unbiasedVLCS}) for Domain Generalization. 1) \textbf{PACS}~\citep{li2017deeperPACS} contains images from four domains, \ieno, 
Photo (P), Art painting (A), Cartoon (C), and Sketch (S). Each domain consists of images in seven object categories. 2) \textbf{Digits-DG} is composed of four digit datasets, including MINIST~\citep{lecun1998gradientMINIST}, MINIST-M~\citep{ganin2015unsupervisedSYN}, SVHN~\citep{netzer2011readingSVHN} and SYN~\citep{ganin2015unsupervisedSYN}. Each dataset is regarded as a domain, which contains ten digit categories from zero to nine. 3) \textbf{Office-Home} is divided into four domains, including Artistic, Clipart, Product and Real World. There are 65 object categories related to the scenes of office and home. 
4) \textbf{VLCS}~\citep{torralba2011unbiasedVLCS} covers images of five object categories from  four domains, \ieno, 
PASCAL VOC 2007, LabelMe, Caltech, and Sun datasets. Following previous DG methods~\citep{li2019episodic,zhou2021domainMixstyle,li2018domainDGMMDAAE,li2018learningMLDG}, we evaluate methods under the leave-one-domain-out protocol, where one domain is used for testing while others for training.

\begin{figure*}[t]
    \centering
    \includegraphics[width=0.98\linewidth]{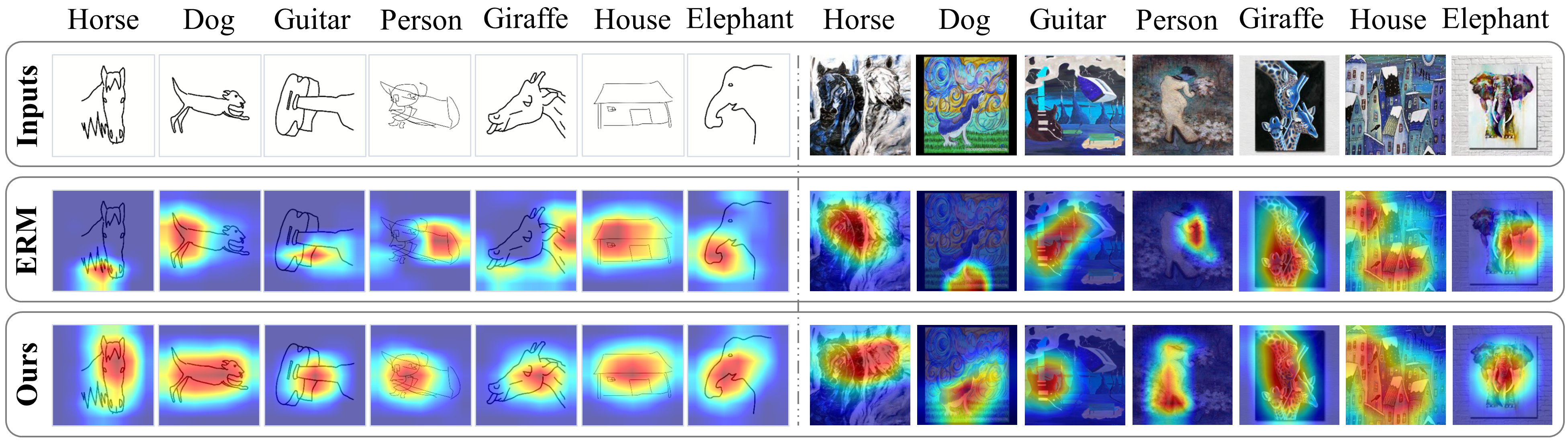}
    \caption{Visualization of the Grad-CAMs w.r.t. classification task for domain generalization. The first row of left/right panels shows the images from art painting (Art)/sketch domains in PACS. The second and third rows show the Grad-CAMs of ERM and our \ours.}
    \label{fig:gradcam}
\end{figure*}

\noindent\textbf{Implementation Details.} 
For PACS and Office-Home, we take ResNet18~\citep{he2016deepResNet} pretrained on ImageNet \citep{deng2009imagenet} as backbone, following \cite{zhou2021domainMixstyle,carlucci2019domain}. We also take the ResNet-50 pretrained on ImageNet as the backbone for PACS, following ~\citet{huang2020selfRSC}.
For VLCS, we take 
AlexNet~\citep{krizhevsky2012imagenetAlexNet} pretrained on ImageNet~\citep{deng2009imagenet}  as our backbone, which is the same as \citep{matsuura2020domainMMLD,dou2019domainMASF}. 
For Digits-DG, we adopt the model architecture used in previous works \citep{zhou2020learningL2A-OT,zhou2021domainMixstyle}. We cluster three clusters within each class in training datasets.
All reported results are averaged among six runs.
More implementation details can be found in the \textbf{Appendix}~\ref{sec:more_implementation}.

\subsection{Ablation Study}
\label{sec: ablation}
\noindent \textbf{Effectiveness of \ours.}
Our proposed \ours~enables models to have superior generalization ability by effectively causal feature learning. We compare our \ours~with the popular 
\begin{wraptable}{r}{0pt}
\setlength{\tabcolsep}{0.5mm}{\begin{tabular}{c|cccc|c}
\hline \hline
Method                        & A                                                    & C                     & P                                                    & S                     & Avg.                                                 \\ \hline 
MatchDG                       & 85.6                                                & 82.1                &  97.9 & 78.8                 & 86.1   \\ 
RSC                           & 87.9                                                & \textbf{82.2}                 & 97.9                                                & 83.4                 & 87.8 \\ 
FACT                          & 89.6 & 81.8                 & 96.8                                                & 84.5                 & 88.2                                                \\ 
Fish                          & -                                                    & -                     & -                                                    & -                     & 85.5                                                 \\ \hline \hline
\multicolumn{1}{c|}{ERM$^*$}      & \multicolumn{1}{c}{88.0}                                & \multicolumn{1}{c}{78.8} & \multicolumn{1}{c}{\textbf{98.2}}                                & \multicolumn{1}{c|}{81.7} & \multicolumn{1}{c}{86.7}                                 \\ 
\rowcolor[gray]{0.9} \multicolumn{1}{c|}{ERM$^*$+\ours} & \textbf{89.7}  & \multicolumn{1}{c}{\textbf{82.2}} & 97.9  & \multicolumn{1}{c|}{\textbf{86.2}} & \multicolumn{1}{c}{\textbf{89.0}}    \\ \hline \hline
\end{tabular}}
\caption{Classification accuracy (\%) of different DG methods on PACS with ResNet-50.}
\label{tab:resnet50}
\vspace{-6mm}
\end{wraptable}
meta-learning strategy MAML, which aims to explore the commonly optimal optimization direction for all tasks (\ieno, different domains), on two baselines. 1) ERM: training models only on source domains with simple data augmentations including flip and translation. 2) ERM$^*$: training models on source domains with
AutoAugment~\citep{cubuk2018autoaugmentAA}. 
The results on PACS are shown in Table \ref{tab:pacs}, with baseline ERM, ERM+\ours ~outperforms ERM by 2.4 \% in accuracy \textit{without known domain labels}, while MAML only achieves the improvement of 0.1\% \textit{with known domain labels}. Moreover, with another baseline, ERM$^*$+\ours~outperforms ERM$^*$ by 3.2\% and 2.0\% on PACS and Digits-DG respectively. However, MAML is not robust for different baselines and does not work for ERM$^*$.

\noindent \textbf{Different ways to estimate global-scope intervening effects.}
To estimate the global-scope intervening effects efficiently and accurately, we propose a \textit{\cluterthensample} strategy. An alternative is the na\"ive random mini-batch sampling. As shown in Fig. \ref{fig:ablation}(a), the classification accuracy increases along with the increased sampling batch-size. This is because increasing batch-size will result in a better estimation of the global-scope intervention with lower SE. 
However, the memory overhead increases drastically simultaneously. 
In contrast, our \textit{\cluterthensample} outperforms the na\"ive random sampling by 0.7\% with lower memory utilization (the batch-size is fixed as 256 for PACS in our case). We also conduct experiments on the influence of the number of clusters $K^\dagger$ for all datasets. The results in Fig. \ref{fig:ablation}(b) show that more clusters result in better performance, and the performance is saturated when $K^\dagger=21$ for PACS (\ieno, three clusters for each class).
Because more clusters lead to lower intra-cluster variance $\sigma_k$ of the $k^{th}$ cluster and more accurate estimation of the global-scope intervention with lower SE, which is derived in the \textbf{Appendix}~\ref{sec:statistical_analysis}. And three clusters for each class are enough for global-scope intervention modeling.

\noindent\textbf{Different choices of $Z$.} \tcb{We explore the effects of different choices of $Z$ from shallow layer features to deep layer features. The experiments and analysis are shown in Appendix~\ref{sec:choicez}, which reveals that shallow features are better.}

\subsection{Comparison with State-of-the-arts}
\label{sec: comparisons}

\noindent\textbf{PACS.} 
As shown in Table \ref{tab:pacs}, our proposed \ours~on top of ERM$^*$ achieves the best performance on PACS, outperforming the SOTA works including MixStyle~\citep{zhou2021domainMixstyle}, RSC~\citep{huang2020selfRSC}, MatchDG~\citep{mahajan2021domainCausalMatch} and FACT~\cite{xu2021fourierFACT}
Moreover, ERM$^*$+\ours~is clearly better than previous gradient-based and meta-learning based methods, \egno,  CrossGrad~\citep{shankar2018generalizingCrossGrad}, MLDG~\citep{li2018learningMLDG}, MetaReg~\citep{balaji2018metareg}, and MAML~\citep{finn2017modelMAML},  thanks to the more accurate global intervening effects modeling in \ours.
On the most challenging domain \textit{sketch}, our ERM$^*$+\ours~outperforms all previous methods by a large margin ($\geq\textbf{2.3\%}$), which demonstrates that \ours ~can learn the causal visual features by removing the influence from confounders efficiently. The experiments on PACS with ResNet-50 are shown in Table~\ref{tab:resnet50}, where our~\ours~is clearly superior to the SOTA methods, \textit{e.g.,} RSC~\citep{huang2020selfRSC}, MatchDG~\citep{mahajan2021domainCausalMatch}, FACT~\citep{xu2021fourierFACT} and recent Fish~\citep{shi2021gradientFish}. 

\noindent\textbf{Digits-DG.}  
As shown in Table. \ref{tab:pacs}, 
our ERM$^*$+\ours~achieves the best performance, outperforming the MAML~\citep{finn2017modelMAML} by 1.4\% and 
the recent MixStyle~\citep{zhou2021domainMixstyle} by 5.2\%. For the most challenging domains (\ieno, MNIST-M and SVHN), ERM$^*$+\ours~improves the accuracy of ERM$^*$ by 3.8\% and 2.5\% respectively.

\noindent\textbf{Office-Home.} 
As shown in Table~\ref{tab: office-home} of Appendix,  
our ERM$^*$+\ours~achieves the best performance of $66.2\%$, exceeding the previous state-of-the-art method MixStyle~\citep{zhou2021domainMixstyle} by 0.7\%. Furthermore, ERM$^*$+\ours~improves ERM$^*$ on the challenging \textit{Clipart} by a margin of 3.2\%.

\noindent\textbf{VLCS.}  
The results on VLCS are shown in Table \ref{tab: vlcs} of Appendix, where
our ERM$^*$+\ours~achieves the state-of-the-art result, outperforming the second best MASF~\citep{dou2019domainMASF} by 0.6\%, which is based on meta-learning. Moreover, \ours~brings the improvement of 1.7\% for ERM$^*$.

\subsection{Feature Visualization}
\label{sec: feat}
To validate that \ours~actually learns the causal visual features,  
we visualize and compare the Grad-CAM~\citep{selvaraju2017gradGradcam} of ERM and ERM+\ours~in Fig. \ref{fig:gradcam}. Intervened by the confounders (\egno, background), ERM easily focuses on the object-irrelevant regions (\ieno, non-causal features), impeding the model's generalization ability. In contrast, thanks to the guidance of \ours, ERM+\ours~is prone to focus more on the foreground object regions (\ieno, causal features). 
Further, we visualize the learned features by t-SNE~\citep{saito2019strongTSNE} on 
Digits-DG in the \textbf{Appendix}~\ref{sec:more_feature_vis}.

\section{Conclusion}
In this paper, 
we propose a novel method dubbed \textit{\ourslong} (\ours) for learning causal visual features without explicit identification and exploitation of confounders. 
Particularly, motivated and based on the front-door criterion, we model the interventions among samples and approximate the global-scope intervening effects for causal visual feature learning.
Extensive experimental results on domain generalization validate that our \ours~can help a model to achieve superior generalization capability by learning causal features, without the need of identifying confounders. 
Our method is generic which should be applicable to other fields such as NLP. We leave this as future work.

\clearpage
%
%
\bibliography{iclr2023_conference}

\begin{thebibliography}{65}
\providecommand{\natexlab}[1]{#1}
\providecommand{\url}[1]{\texttt{#1}}
\expandafter\ifx\csname urlstyle\endcsname\relax
  \providecommand{\doi}[1]{doi: #1}\else
  \providecommand{\doi}{doi: \begingroup \urlstyle{rm}\Url}\fi

\bibitem[Antol et~al.(2015)Antol, Agrawal, Lu, Mitchell, Batra, Zitnick, and
  Parikh]{antol2015VQA}
Stanislaw Antol, Aishwarya Agrawal, Jiasen Lu, Margaret Mitchell, Dhruv Batra,
  C~Lawrence Zitnick, and Devi Parikh.
\newblock Vqa: Visual question answering.
\newblock In \emph{ICCV}, pp.\  2425--2433, 2015.

\bibitem[Balaji et~al.(2018)Balaji, Sankaranarayanan, and
  Chellappa]{balaji2018metareg}
Yogesh Balaji, Swami Sankaranarayanan, and Rama Chellappa.
\newblock Metareg: Towards domain generalization using meta-regularization.
\newblock In \emph{NeurIPS}, volume~31, pp.\  998--1008, 2018.

\bibitem[Carlucci et~al.(2019)Carlucci, D'Innocente, Bucci, Caputo, and
  Tommasi]{carlucci2019domain}
Fabio~M Carlucci, Antonio D'Innocente, Silvia Bucci, Barbara Caputo, and
  Tatiana Tommasi.
\newblock Domain generalization by solving jigsaw puzzles.
\newblock In \emph{CVPR}, pp.\  2229--2238, 2019.

\bibitem[Cha et~al.(2021)Cha, Cho, Lee, Park, Lee, and Park]{cha2021domain}
Junbum Cha, Hancheol Cho, Kyungjae Lee, Seunghyun Park, Yunsung Lee, and
  Sungrae Park.
\newblock Domain generalization needs stochastic weight averaging for
  robustness on domain shifts.
\newblock \emph{arXiv preprint arXiv:2102.08604}, 2021.

\bibitem[Cubuk et~al.(2018)Cubuk, Zoph, Mane, Vasudevan, and
  Le]{cubuk2018autoaugmentAA}
Ekin~D Cubuk, Barret Zoph, Dandelion Mane, Vijay Vasudevan, and Quoc~V Le.
\newblock Autoaugment: Learning augmentation policies from data.
\newblock In \emph{CVPR}, 2018.

\bibitem[Deng et~al.(2009)Deng, Dong, Socher, Li, Li, and
  Fei-Fei]{deng2009imagenet}
Jia Deng, Wei Dong, Richard Socher, Li-Jia Li, Kai Li, and Li~Fei-Fei.
\newblock Imagenet: A large-scale hierarchical image database.
\newblock In \emph{CVPR}, pp.\  248--255. Ieee, 2009.

\bibitem[Dou et~al.(2019)Dou, Coelho~de Castro, Kamnitsas, and
  Glocker]{dou2019domainMASF}
Qi~Dou, Daniel Coelho~de Castro, Konstantinos Kamnitsas, and Ben Glocker.
\newblock Domain generalization via model-agnostic learning of semantic
  features.
\newblock In \emph{NeurIPS}, volume~32, pp.\  6450--6461, 2019.

\bibitem[Finn et~al.(2017)Finn, Abbeel, and Levine]{finn2017modelMAML}
Chelsea Finn, Pieter Abbeel, and Sergey Levine.
\newblock Model-agnostic meta-learning for fast adaptation of deep networks.
\newblock In \emph{ICML}, pp.\  1126--1135. PMLR, 2017.

\bibitem[Ganin \& Lempitsky(2015)Ganin and Lempitsky]{ganin2015unsupervisedSYN}
Yaroslav Ganin and Victor Lempitsky.
\newblock Unsupervised domain adaptation by backpropagation.
\newblock In \emph{ICML}, pp.\  1180--1189. PMLR, 2015.

\bibitem[Ghosh(2002)]{ghosh2002elementsSamplingTheory}
Subir Ghosh.
\newblock Elements of sampling theory and methods, 2002.

\bibitem[He et~al.(2016)He, Zhang, Ren, and Sun]{he2016deepResNet}
Kaiming He, Xiangyu Zhang, Shaoqing Ren, and Jian Sun.
\newblock Deep residual learning for image recognition.
\newblock In \emph{CVPR}, pp.\  770--778, 2016.

\bibitem[He et~al.(2017)He, Gkioxari, Doll{\'a}r, and Girshick]{he2017maskRCNN}
Kaiming He, Georgia Gkioxari, Piotr Doll{\'a}r, and Ross Girshick.
\newblock Mask r-cnn.
\newblock In \emph{ICCV}, pp.\  2961--2969, 2017.

\bibitem[Hu et~al.(2021)Hu, Tang, Miao, Hua, and Zhang]{hu2021distilling}
Xinting Hu, Kaihua Tang, Chunyan Miao, Xian-Sheng Hua, and Hanwang Zhang.
\newblock Distilling causal effect of data in class-incremental learning.
\newblock In \emph{CVPR}, pp.\  3957--3966, 2021.

\bibitem[Huang et~al.(2020)Huang, Wang, Xing, and Huang]{huang2020selfRSC}
Zeyi Huang, Haohan Wang, Eric~P Xing, and Dong Huang.
\newblock Self-challenging improves cross-domain generalization.
\newblock In \emph{ECCV}, pp.\  124--140, 2020.

\bibitem[Imbens \& Rubin(2015)Imbens and Rubin]{imbens2015causal}
Guido~W Imbens and Donald~B Rubin.
\newblock \emph{Causal inference in statistics, social, and biomedical
  sciences}.
\newblock Cambridge University Press, 2015.

\bibitem[Jin et~al.(2020)Jin, Lan, Zeng, and Chen]{jin2020featureFAR}
Xin Jin, Cuiling Lan, Wenjun Zeng, and Zhibo Chen.
\newblock Feature alignment and restoration for domain generalization and
  adaptation.
\newblock \emph{arXiv preprint arXiv:2006.12009}, 2020.

\bibitem[Krizhevsky et~al.(2012)Krizhevsky, Sutskever, and
  Hinton]{krizhevsky2012imagenetAlexNet}
Alex Krizhevsky, Ilya Sutskever, and Geoffrey~E Hinton.
\newblock Imagenet classification with deep convolutional neural networks.
\newblock In \emph{NeurIPS}, volume~25, pp.\  1097--1105, 2012.

\bibitem[LeCun et~al.(1998)LeCun, Bottou, Bengio, and
  Haffner]{lecun1998gradientMINIST}
Yann LeCun, L{\'e}on Bottou, Yoshua Bengio, and Patrick Haffner.
\newblock Gradient-based learning applied to document recognition.
\newblock \emph{Proceedings of the IEEE}, 86\penalty0 (11):\penalty0
  2278--2324, 1998.

\bibitem[Li et~al.(2017)Li, Yang, Song, and Hospedales]{li2017deeperPACS}
Da~Li, Yongxin Yang, Yi-Zhe Song, and Timothy~M Hospedales.
\newblock Deeper, broader and artier domain generalization.
\newblock In \emph{ICCV}, pp.\  5542--5550, 2017.

\bibitem[Li et~al.(2018{\natexlab{a}})Li, Yang, Song, and
  Hospedales]{li2018learningMLDG}
Da~Li, Yongxin Yang, Yi-Zhe Song, and Timothy~M Hospedales.
\newblock Learning to generalize: Meta-learning for domain generalization.
\newblock In \emph{AAAI}, 2018{\natexlab{a}}.

\bibitem[Li et~al.(2019)Li, Zhang, Yang, Liu, Song, and
  Hospedales]{li2019episodic}
Da~Li, Jianshu Zhang, Yongxin Yang, Cong Liu, Yi-Zhe Song, and Timothy~M
  Hospedales.
\newblock Episodic training for domain generalization.
\newblock In \emph{ICCV}, pp.\  1446--1455, 2019.

\bibitem[Li et~al.(2020)Li, Yang, Song, and Hospedales]{li2020sequentialSMLDG}
Da~Li, Yongxin Yang, Yi-Zhe Song, and Timothy Hospedales.
\newblock Sequential learning for domain generalization.
\newblock In \emph{ECCV}, pp.\  603--619. Springer, 2020.

\bibitem[Li et~al.(2018{\natexlab{b}})Li, Pan, Wang, and
  Kot]{li2018domainDGMMDAAE}
Haoliang Li, Sinno~Jialin Pan, Shiqi Wang, and Alex~C Kot.
\newblock Domain generalization with adversarial feature learning.
\newblock In \emph{CVPR}, pp.\  5400--5409, 2018{\natexlab{b}}.

\bibitem[Li et~al.(2018{\natexlab{c}})Li, Tian, Gong, Liu, Liu, Zhang, and
  Tao]{li2018deepDeepConditionIDN}
Ya~Li, Xinmei Tian, Mingming Gong, Yajing Liu, Tongliang Liu, Kun Zhang, and
  Dacheng Tao.
\newblock Deep domain generalization via conditional invariant adversarial
  networks.
\newblock In \emph{ECCV}, pp.\  624--639, 2018{\natexlab{c}}.

\bibitem[Liu et~al.(2020)Liu, Dou, and Heng]{liu2020shapeShapeMeta}
Quande Liu, Qi~Dou, and Pheng-Ann Heng.
\newblock Shape-aware meta-learning for generalizing prostate mri segmentation
  to unseen domains.
\newblock In \emph{International Conference on Medical Image Computing and
  Computer-Assisted Intervention}, pp.\  475--485. Springer, 2020.

\bibitem[Liu et~al.(2021)Liu, Lin, Cao, Hu, Wei, Zhang, Lin, and
  Guo]{liu2021swin}
Ze~Liu, Yutong Lin, Yue Cao, Han Hu, Yixuan Wei, Zheng Zhang, Stephen Lin, and
  Baining Guo.
\newblock Swin transformer: Hierarchical vision transformer using shifted
  windows.
\newblock In \emph{ICCV}, 2021.

\bibitem[Mahajan et~al.(2021{\natexlab{a}})Mahajan, Tople, and
  Sharma]{mahajan2021domainCausalMatch}
Divyat Mahajan, Shruti Tople, and Amit Sharma.
\newblock Domain generalization using causal matching.
\newblock In \emph{International Conference on Machine Learning}, pp.\
  7313--7324. PMLR, 2021{\natexlab{a}}.

\bibitem[Mahajan et~al.(2021{\natexlab{b}})Mahajan, Tople, and
  Sharma]{mahajan2021domainDGCausalMatch}
Divyat Mahajan, Shruti Tople, and Amit Sharma.
\newblock Domain generalization using causal matching.
\newblock In \emph{ICML}, pp.\  7313--7324. PMLR, 2021{\natexlab{b}}.

\bibitem[Matsuura \& Harada(2020)Matsuura and Harada]{matsuura2020domainMMLD}
Toshihiko Matsuura and Tatsuya Harada.
\newblock Domain generalization using a mixture of multiple latent domains.
\newblock In \emph{AAAI}, volume~34, pp.\  11749--11756, 2020.

\bibitem[Motiian et~al.(2017)Motiian, Piccirilli, Adjeroh, and
  Doretto]{motiian2017unifiedCCSA}
Saeid Motiian, Marco Piccirilli, Donald~A Adjeroh, and Gianfranco Doretto.
\newblock Unified deep supervised domain adaptation and generalization.
\newblock In \emph{ICCV}, pp.\  5715--5725, 2017.

\bibitem[Muandet et~al.(2013)Muandet, Balduzzi, and
  Sch{\"o}lkopf]{muandet2013domainFIRSTDG}
Krikamol Muandet, David Balduzzi, and Bernhard Sch{\"o}lkopf.
\newblock Domain generalization via invariant feature representation.
\newblock In \emph{ICML}, pp.\  10--18, 2013.

\bibitem[Murnane \& Willett(2010)Murnane and Willett]{murnane2010methods}
Richard~J Murnane and John~B Willett.
\newblock \emph{Methods matter: Improving causal inference in educational and
  social science research}.
\newblock Oxford University Press, 2010.

\bibitem[Netzer et~al.(2011)Netzer, Wang, Coates, Bissacco, Wu, and
  Ng]{netzer2011readingSVHN}
Yuval Netzer, Tao Wang, Adam Coates, Alessandro Bissacco, Bo~Wu, and Andrew~Y
  Ng.
\newblock Reading digits in natural images with unsupervised feature learning.
\newblock 2011.

\bibitem[Pearl(2009{\natexlab{a}})]{pearl2009causal}
Judea Pearl.
\newblock Causal inference in statistics: An overview.
\newblock \emph{Statistics surveys}, 3:\penalty0 96--146, 2009{\natexlab{a}}.

\bibitem[Pearl(2009{\natexlab{b}})]{pearl2009causality}
Judea Pearl.
\newblock \emph{Causality}.
\newblock Cambridge university press, 2009{\natexlab{b}}.

\bibitem[Pearl et~al.(2016)Pearl, Glymour, and Jewell]{pearl2016causal}
Judea Pearl, Madelyn Glymour, and Nicholas~P Jewell.
\newblock \emph{Causal inference in statistics: A primer}.
\newblock John Wiley \& Sons, 2016.

\bibitem[Pelleg et~al.(2000)Pelleg, Moore, et~al.]{pelleg2000xKmeans}
Dan Pelleg, Andrew~W Moore, et~al.
\newblock X-means: Extending k-means with efficient estimation of the number of
  clusters.
\newblock In \emph{ICML}, volume~1, pp.\  727--734, 2000.

\bibitem[Qiao et~al.(2020)Qiao, Zhao, and Peng]{qiao2020learningMADA}
Fengchun Qiao, Long Zhao, and Xi~Peng.
\newblock Learning to learn single domain generalization.
\newblock In \emph{CVPR}, pp.\  12556--12565, 2020.

\bibitem[Redmon et~al.(2016)Redmon, Divvala, Girshick, and
  Farhadi]{redmon2016youYOLO}
Joseph Redmon, Santosh Divvala, Ross Girshick, and Ali Farhadi.
\newblock You only look once: Unified, real-time object detection.
\newblock In \emph{CVPR}, pp.\  779--788, 2016.

\bibitem[Rubin(1986)]{rubin1986statistics}
Donald~B Rubin.
\newblock Statistics and causal inference: Comment: Which ifs have causal
  answers.
\newblock \emph{Journal of the American Statistical Association}, 81\penalty0
  (396):\penalty0 961--962, 1986.

\bibitem[Saito et~al.(2019)Saito, Ushiku, Harada, and
  Saenko]{saito2019strongTSNE}
Kuniaki Saito, Yoshitaka Ushiku, Tatsuya Harada, and Kate Saenko.
\newblock Strong-weak distribution alignment for adaptive object detection.
\newblock In \emph{CVPR}, pp.\  6956--6965, 2019.

\bibitem[Selvaraju et~al.(2017)Selvaraju, Cogswell, Das, Vedantam, Parikh, and
  Batra]{selvaraju2017gradGradcam}
Ramprasaath~R Selvaraju, Michael Cogswell, Abhishek Das, Ramakrishna Vedantam,
  Devi Parikh, and Dhruv Batra.
\newblock Grad-cam: Visual explanations from deep networks via gradient-based
  localization.
\newblock In \emph{ICCV}, pp.\  618--626, 2017.

\bibitem[Seo et~al.(2020)Seo, Suh, Kim, Kim, Han, and
  Han]{seo2020learningDomainSpecificNorm}
Seonguk Seo, Yumin Suh, Dongwan Kim, Geeho Kim, Jongwoo Han, and Bohyung Han.
\newblock Learning to optimize domain specific normalization for domain
  generalization.
\newblock In \emph{ECCV}, pp.\  68--83. Springer, 2020.

\bibitem[Shankar et~al.(2018)Shankar, Piratla, Chakrabarti, Chaudhuri, Jyothi,
  and Sarawagi]{shankar2018generalizingCrossGrad}
Shiv Shankar, Vihari Piratla, Soumen Chakrabarti, Siddhartha Chaudhuri, Preethi
  Jyothi, and Sunita Sarawagi.
\newblock Generalizing across domains via cross-gradient training.
\newblock In \emph{ICLR}, 2018.

\bibitem[Shen et~al.(2021)Shen, Liu, He, Zhang, Xu, Yu, and
  Cui]{shen2021towardsSurveyCUIPENG}
Zheyan Shen, Jiashuo Liu, Yue He, Xingxuan Zhang, Renzhe Xu, Han Yu, and Peng
  Cui.
\newblock Towards out-of-distribution generalization: A survey.
\newblock \emph{arXiv preprint arXiv:2108.13624}, 2021.

\bibitem[Shi et~al.(2021)Shi, Seely, Torr, Siddharth, Hannun, Usunier, and
  Synnaeve]{shi2021gradientFish}
Yuge Shi, Jeffrey Seely, Philip~HS Torr, N~Siddharth, Awni Hannun, Nicolas
  Usunier, and Gabriel Synnaeve.
\newblock Gradient matching for domain generalization.
\newblock \emph{arXiv preprint arXiv:2104.09937}, 2021.

\bibitem[Sun et~al.(2016)Sun, Feng, and Saenko]{sun2016returnCORAL}
Baochen Sun, Jiashi Feng, and Kate Saenko.
\newblock Return of frustratingly easy domain adaptation.
\newblock In \emph{AAAI}, 2016.

\bibitem[Taori et~al.(2020)Taori, Dave, Shankar, Carlini, Recht, and
  Schmidt]{taori2020measuring}
Rohan Taori, Achal Dave, Vaishaal Shankar, Nicholas Carlini, Benjamin Recht,
  and Ludwig Schmidt.
\newblock Measuring robustness to natural distribution shifts in image
  classification.
\newblock In \emph{NeurIPS}, 2020.

\bibitem[Torralba \& Efros(2011)Torralba and Efros]{torralba2011unbiasedVLCS}
Antonio Torralba and Alexei~A Efros.
\newblock Unbiased look at dataset bias.
\newblock In \emph{CVPR}, pp.\  1521--1528. IEEE, 2011.

\bibitem[Venkateswara et~al.(2017)Venkateswara, Eusebio, Chakraborty, and
  Panchanathan]{venkateswara2017deepOffice-Home}
Hemanth Venkateswara, Jose Eusebio, Shayok Chakraborty, and Sethuraman
  Panchanathan.
\newblock Deep hashing network for unsupervised domain adaptation.
\newblock In \emph{CVPR}, pp.\  5018--5027, 2017.

\bibitem[Wang et~al.(2021)Wang, Lan, Liu, Ouyang, Zeng, and
  Qin]{wang2021generalizingDGSurvey}
Jindong Wang, Cuiling Lan, Chang Liu, Yidong Ouyang, Wenjun Zeng, and Tao Qin.
\newblock Generalizing to unseen domains: A survey on domain generalization.
\newblock In \emph{IJCAI}, 2021.

\bibitem[Wang et~al.(2020{\natexlab{a}})Wang, Huang, Zhang, and
  Sun]{wang2020commonsense}
Tan Wang, Jianqiang Huang, Hanwang Zhang, and Qianru Sun.
\newblock Visual commonsense r-cnn.
\newblock In \emph{CVPR}, pp.\  10760--10770, 2020{\natexlab{a}}.

\bibitem[Wang et~al.(2020{\natexlab{b}})Wang, Huang, Zhang, and
  Sun]{wang2020visual}
Tan Wang, Jianqiang Huang, Hanwang Zhang, and Qianru Sun.
\newblock Visual commonsense representation learning via causal inference.
\newblock In \emph{CVPR Workshops}, pp.\  378--379, 2020{\natexlab{b}}.

\bibitem[Wei et~al.(2021)Wei, Lan, Zeng, and Chen]{wei2021metaalign}
Guoqiang Wei, Cuiling Lan, Wenjun Zeng, and Zhibo Chen.
\newblock Metaalign: Coordinating domain alignment and classification for
  unsupervised domain adaptation.
\newblock In \emph{CVPR}, pp.\  16643--16653, 2021.

\bibitem[Xu et~al.(2021)Xu, Zhang, Zhang, Wang, and Tian]{xu2021fourierFACT}
Qinwei Xu, Ruipeng Zhang, Ya~Zhang, Yanfeng Wang, and Qi~Tian.
\newblock A fourier-based framework for domain generalization.
\newblock In \emph{Proceedings of the IEEE/CVF Conference on Computer Vision
  and Pattern Recognition}, pp.\  14383--14392, 2021.

\bibitem[Yang et~al.(2021)Yang, Zhang, Qi, and Cai]{yang2021causal}
Xu~Yang, Hanwang Zhang, Guojun Qi, and Jianfei Cai.
\newblock Causal attention for vision-language tasks.
\newblock In \emph{CVPR}, pp.\  9847--9857, 2021.

\bibitem[Yue et~al.(2020)Yue, Zhang, Sun, and Hua]{yue2020interventional}
Zhongqi Yue, Hanwang Zhang, Qianru Sun, and Xian-Sheng Hua.
\newblock Interventional few-shot learning.
\newblock In \emph{NeurIPS}, 2020.

\bibitem[Yue et~al.(2021)Yue, Sun, Hua, and Zhang]{yue2021transporting}
Zhongqi Yue, Qianru Sun, Xian-Sheng Hua, and Hanwang Zhang.
\newblock Transporting causal mechanisms for unsupervised domain adaptation.
\newblock In \emph{ICCV}, pp.\  8599--8608, 2021.

\bibitem[Zhang et~al.(2020)Zhang, Zhang, Tang, Hua, and Sun]{zhang2020causal}
Dong Zhang, Hanwang Zhang, Jinhui Tang, Xiansheng Hua, and Qianru Sun.
\newblock Causal intervention for weakly-supervised semantic segmentation.
\newblock In \emph{NeurIPS}, 2020.

\bibitem[Zhao et~al.(2021)Zhao, Zhong, Yang, Luo, Lin, Li, and
  Sebe]{zhao2021learningMetaReid}
Yuyang Zhao, Zhun Zhong, Fengxiang Yang, Zhiming Luo, Yaojin Lin, Shaozi Li,
  and Nicu Sebe.
\newblock Learning to generalize unseen domains via memory-based multi-source
  meta-learning for person re-identification.
\newblock In \emph{CVPR}, pp.\  6277--6286, 2021.

\bibitem[Zhou et~al.(2020{\natexlab{a}})Zhou, Yang, Hospedales, and
  Xiang]{zhou2020deep}
Kaiyang Zhou, Yongxin Yang, Timothy Hospedales, and Tao Xiang.
\newblock Deep domain-adversarial image generation for domain generalisation.
\newblock In \emph{AAAI}, volume~34, pp.\  13025--13032, 2020{\natexlab{a}}.

\bibitem[Zhou et~al.(2020{\natexlab{b}})Zhou, Yang, Hospedales, and
  Xiang]{zhou2020learningL2A-OT}
Kaiyang Zhou, Yongxin Yang, Timothy Hospedales, and Tao Xiang.
\newblock Learning to generate novel domains for domain generalization.
\newblock In \emph{ECCV}, pp.\  561--578. Springer, 2020{\natexlab{b}}.

\bibitem[Zhou et~al.(2021{\natexlab{a}})Zhou, Liu, Qiao, Xiang, and
  Loy]{zhou2021domainDGSurveyzhou}
Kaiyang Zhou, Ziwei Liu, Yu~Qiao, Tao Xiang, and Chen~Change Loy.
\newblock Domain generalization: A survey.
\newblock \emph{arXiv preprint arXiv:2103.02503}, 2021{\natexlab{a}}.

\bibitem[Zhou et~al.(2021{\natexlab{b}})Zhou, Yang, Qiao, and
  Xiang]{zhou2021domainDGensemble}
Kaiyang Zhou, Yongxin Yang, Yu~Qiao, and Tao Xiang.
\newblock Domain adaptive ensemble learning.
\newblock \emph{IEEE Transactions on Image Processing}, 30:\penalty0
  8008--8018, 2021{\natexlab{b}}.

\bibitem[Zhou et~al.(2021{\natexlab{c}})Zhou, Yang, Qiao, and
  Xiang]{zhou2021domainMixstyle}
Kaiyang Zhou, Yongxin Yang, Yu~Qiao, and Tao Xiang.
\newblock Domain generalization with mixstyle.
\newblock In \emph{ICLR}, 2021{\natexlab{c}}.

\end{thebibliography}
\bibliographystyle{iclr2023_conference}
\clearpage
\appendix
\section{Appendix}
\subsection{More Details on \ours}
\label{sec:more_details}
\subsubsection{Statistical Analysis on Sampling Algorithms}
\label{sec:statistical_analysis}

In this section, we provide the statistical analysis for the superiority of our proposed sampling strategy, \ieno, \textit{\cluterthensample}, by comparing it to the \textit{random mini-batch sampling} strategy. As a result,
we theoretically derive that using our \textit{clustering-then-sampling} has a significant smaller standard error (SE)  in estimating the global-scope gradient, compared with using the \textit{random mini-batch sampling} strategy.

Given a training dataset $D={(x_i, y_i)}_{i=1}^N$, the global-scope gradient can be computed as:
\begin{equation}
    g_\dagger = \sum_{i=1}^{N}g_iP(x_i), where\; P(x_i)=\frac{1}{N} 
    \label{eq:global-scope-gradient}
\end{equation}
However, it is intractable to compute $g_\dagger$ with Eq. \ref{eq:global-scope-gradient} directly, which requires 
traversing over all training data. 

Our \textit{clustering-then-sampling} aims to simulate the global-scope gradient with partial data from the training dataset. Another alternative na\"ive method is \textit{random mini-batch sampling.} 

\noindent
\textbf{\textit{Random mini-batch sampling.}}
With this strategy, the gradient $g_{random}$ of each iteration is computed over a mini-batch data ${\{x_j^r, y_j^r\}}_{j=1}^M$, where the mini-batch data with size $M$ is randomly sampled from the training data:
\begin{equation}
    g_{random}=\sum_{j=1}^{M}g_jP(x_j^r), where\; P(x_j^r)=\frac{1}{M}
    \label{eq:randomsampling}
\end{equation}
According to the sampling theory~\citep{ghosh2002elementsSamplingTheory}, the expectation of $g_{random}$ can be represented as: 
\begin{equation}
    E(g_{random})=E(\frac{1}{M}\sum_{j=1}^{M}g_j)=\frac{1}{M}\sum_{j=1}^{M}E(g_j)=\mu,
\end{equation}
where $\mu$ denotes the expectation 
of gradients of all samples in the training data.
To measure this strategy's estimation accuracy of approximating the global-scope gradient $g_{random}$ in Eq. \ref{eq:global-scope-gradient}, we can derive its SE as:
\begin{equation}
    SE_{random} = \frac{\sigma^2}{M}(1-\frac{M-1}{N-1}),
    \label{eq:4}
\end{equation}
where $\sigma^2$ is the variance of the gradients of all training samples. Considering $M\ll N$, Eq. \ref{eq:4} can be further approximated as:
\begin{equation}
    SE_{random} \approx \frac{\sigma^2}{M}.
    \label{eq:5}
\end{equation}

\noindent
\textbf{\textit{\cluterthensample}.}
In contrast, in our \cluterthensample, we
first cluster the training samples of each class in the dataset into $K$ cluster with $K$-means algorithm~\citep{pelleg2000xKmeans} and totally obtain $K^\dagger$ clusters for the whole training data. 
We denote the mean and variance of the gradients of the $k^{th}$ cluster $K_k$ as $\mu_k$ and $\sigma_k^2$ respectively. Since the clustering, the variance of the sample distribution in each cluster is significantly smaller than the variance of the sample distribution in whole training data. Therefore, the gradient variance of the $k^{th}$ cluster $\sigma_k^2$ is smaller than $\sigma^2$.
Then we sample $N_k$ samples from the $k^{th}$ cluster
to form a mini-batch with the size of $M$ (\ieno, $M=\sum_{k=1}^{K^\dagger} N_k$). Here, $N_k$ is 
proportional to the ratio $P(K_k)$ (\ieno, $N_k=MP(K_k)$) and $P(K_k)$ is ratio of the size  of the $k^{th}$ cluster $N^K_k$  to the size of entire training data $N$ (\ieno, $P(K_k)=\frac{N^K_k}{N}$).
 Then the global-scope gradient can be computed with:
\begin{equation}
    g_{ours} = \frac{1}{M}\sum_{k=1}^{K^\dagger}\sum_{j=1}^{N_k}g_{{\Tilde{x}_{j,k}} \in K_k} 
    \label{eq:6}
\end{equation}
Based on stratified sampling theory \citep{ghosh2002elementsSamplingTheory}, the expectation of $g_{ours}$ can be computed as:
\begin{equation}
    E(g_{ours})=\sum_{k=1}^{K^\dagger}\frac{N_k}{M}\mu_k = \sum_{k=1}^{K^\dagger}\frac{MP(K_k)}{M}\mu_k=\mu
    \label{eq:7}
\end{equation}

The standard error (SE) of our \textit{clustering-then-sampling} can be derived as :
\begin{equation}
    SE_{ours}=\sum_{k=1}^{K^\dagger}P(K_k)^2\frac{1}{N_k}(1-\frac{N_k-1}{N_k^K-1})\sigma_k^2,
    \label{eq:9}
\end{equation}
where $N_k^K$ denotes the number of all samples in the $k^{th}$ cluster. 
Considering $N_k\ll N_k^K$, 
\begin{equation}
    SE_{ours}\approx\sum_{k=1}^{K^\dagger}P(K_k)^2\frac{1}{N_k}\sigma_k^2=\sum_{k=1}^{K^\dagger}\frac{P(K_k)}{M}\sigma_k^2
    \label{eq:10}
\end{equation}
In 
clustering, the intra-cluster gradient variance $\sigma_k^2$ reduces with the increasing the number of clusters. Thus, we can get  $\sigma_k^2<\sigma^2$ when $1\le k\le K^\dagger$. We represent the maximum value of $\sigma^2_k$ as $(\sigma^2_k)_{max}=max\{\sigma^2_k| 1\le k \le K^\dagger\}$. The Eq. \ref{eq:10} can be rewritten as:
\begin{align}
    SE_{ours}&=\sum_{k=1}^{K^\dagger}\frac{P(K_k)}{M}\sigma_k^2 \le ({\sigma_k^2})_{max}\sum_{k=1}^{K^\dagger}\frac{P(K_k)}{M} \notag\\ 
    &= \frac{1}{M}(\sigma_k^2)_{max} < \frac{1}{M} \sigma^2 = SE_{random}
    \label{eq:11}
\end{align}
Based on the Eq. \ref{eq:11}, we can draw a conclusion that our \textit{clustering-then-sampling} is better than \textit{random mini-batch sampling} for simulating the global-scope gradient accurately. Furthermore, from the Eq.~\ref{eq:11}, we can let $(\sigma_k^2)_{max} \ll \sigma^2$ by increasing the number of clusters, and obtain the $SE_{ours} \ll SE_{random}$. 
\begin{algorithm}[t]
\caption{\ourslong}
\label{alg:cicf}
\begin{algorithmic}[1]
\State \textbf{Input:} Training dataset $\{(x_i, y_i)\}_{i=1}^{N}$. 
\State \textbf{Init:}
 learning rate: $\alpha$, $\beta$; model $f$ with parameter $\theta$.
\State Obtain $K^\dagger$ clusters from training data by clustering the samples of each class into $K$ clusters.
\While{not converge}
\State Sample $M$ samples from $K^\dagger$ clusters as a batch. 
\State Estimate global intervention with $g_\dagger$.  \Comment{Eq. \ref{equ:cluster_our}}
\State Update $f$ with $g_\dagger$ as: $\theta_\dagger = \theta-\alpha g_\dagger$.         
\State Compute the loss $\mathcal{L}_{\ours}$. \Comment{Eq.~\ref{equ:celoss}}
\State Update $\theta \gets \theta -\beta \nabla_\theta  \mathcal{L}_{\ours}$.
\EndWhile
\end{algorithmic}
\end{algorithm}
\subsubsection{Experimental Evidence for the Significant Difference between two Sampling Strategies.}
To further demonstrate the difference between our \textit{\cluterthensample} and \textit{random mini-batch sampling}, we introduce a metric $E=\sum_{k=1}^{K^\dagger}(|N_k-R_k|)$ to measure the difference degree between two sampling strategies, where $N_k$ and $R_k$ denote the number of sampled instances from the $k^{th}$ cluster using \textit{\cluterthensample} and \textit{random mini-batch sampling}, respectively. We conduct two experiments on PACS with batch-size $M=256$ as follows. 1) We cluster three clusters for each of the seven classes in PACS and totally obtain 21 clusters. The average E is 55, \ieno, the difference ratio $E/M$  between \textit{\cluterthensample} and \textit{random mini-batch sampling}  is  $55/256=21.5\%$. 2) We consider the class prior for \textit{random mini-batch sampling}, and adopt the \textit{random mini-batch sampling} weighted by the number of each class. We have $E$ as 46 and the difference ratio $E/M=46/256=18.0\%$. 
Based on the above experiments, we can find that our \textit{\cluterthensample} is significantly different from \textit{random mini-batch sampling}.

\subsubsection{More Details on the Setting of Sampling Algorithm}
Our \ours~aims to simulate the global intervening effects from the perspective of  optimization with:
\begin{equation}
    g_\dagger= \frac{1}{M}\sum_{k=1}^{K^\dagger}\sum_{j=1}^{N_k} g_{\Tilde{x}_{j,k}\in K_k},
\end{equation}
where a mini-batch sampling procedure is applied to $\Tilde{x}_{j,k} \in K_k$ (\ieno, clustering-then-sampling). After obtaining the updated $\theta_\dagger$ with $\theta_\dagger=\theta-\alpha*g_\dagger$, we can obtain the loss function $\mathcal{L}_\ours$ and get the reliable optimization direction against the intervening effects of confounders:
\begin{equation}
    \mathcal{L}_\ours = \sum_{x\in X}\mathcal{L}_{ce}(f_{\theta_\dagger}(h(x)), y), 
\end{equation}
where another mini-batch sampling procedure is applied to the variable $x$. 

As above, there are two mini-batch sampling in our \ours~adopted for computing global-scope gradient and loss function $L_\ours$ respectively. They are denoted by $\{\Tilde{x},\Tilde{y}\}_{i=1}^{M}$ and $\{x, y\}_{i=1}^{M^l}$, respectively.
To clarify the composition of the samples in each mini-batch, we first define some notations, respectively as follows:
\begin{itemize}
    \item $N$: The number of all samples in the training dataset
    \item $N_k^K$: The number of all samples in the $k^{th}$ cluster.
    \item $N_k$: The number of sampled samples in the $k^{th}$ cluster to form a mini-batch.
    \item $M$: The number of all samples in a mini-batch for computing global-scope gradient.
    \item $M^l$: The number of all samples in a mini-batch for computing $L_\ours$.
    \item $P(K_k)$: The ratio of all samples $N_k^K$ in the $k^{th}$ cluster to all training data $N$, which is represented as $P(K_k)=\frac{N_k^K}{N}$
\end{itemize}

For computing global-scope gradient, we respectively sample $N_k$ samples from the $k^{th}$ cluster to form a mini-batch $\{\Tilde{x},\Tilde{y}\}_{i=1}^{M}$. 
$N_k$ can be set with two strategies, respectively $N_k=MP(K_k)$ and $N_k=\frac{M}{K^\dagger}$, responding to two scheme for computing $L_\ours$. 

When we directly consider the unbalance question between different clusters in computing $L_\ours$, we can sample $\frac{M^l}{K^\dagger}$ samples from each cluster to form a mini-batch $\{x, y\}_{i=1}^{M^l}$, and corresponding $N_k$ is $N_k=\frac{M}{K^\dagger}$. On the contrary, when we randomly sample $M^l$ samples from all training data to compute $L_\ours$, the corresponding $N_k$ is $N_k=MP(K_k)$, which needs to model the intervention caused by the unbalance question. 

\subsection{More Details on Back-door/Front-door criterion}
\label{sec:more_backfront}

Back-door and front-door criteria have been proposed in~\citet{pearl2016causal,pearl2009causal} to reveal the causality between two variables $X$ and $Y$. We further clarify their basics mathematically in this section.

\begin{figure}[t]
    \centering
    \includegraphics[width=0.6\linewidth]{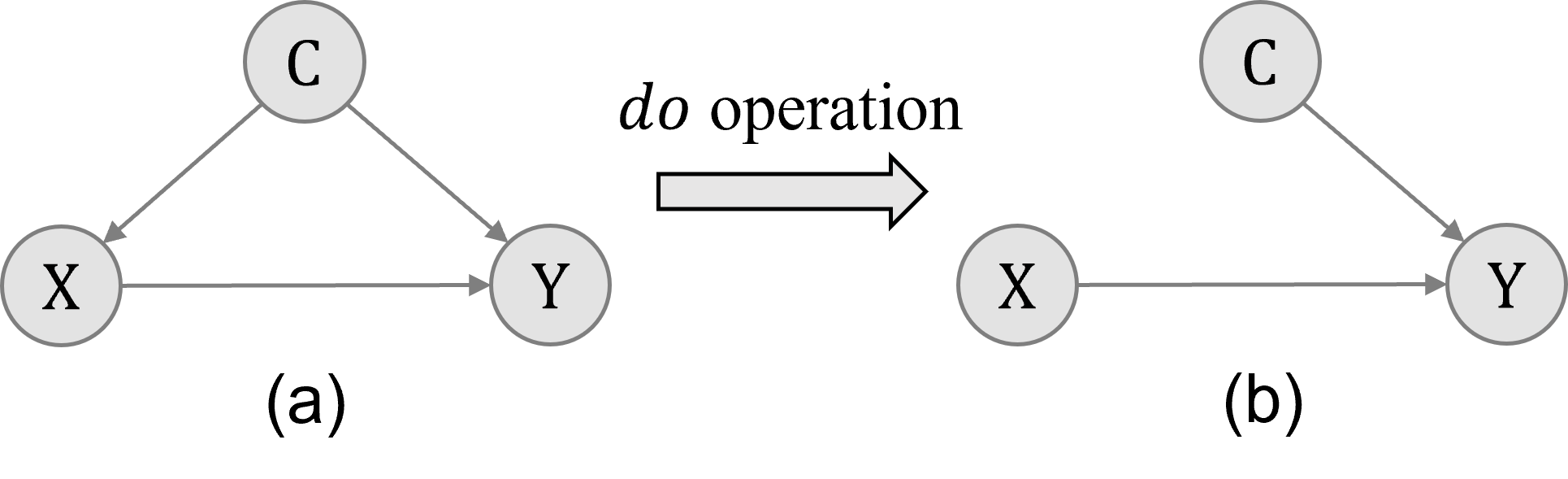}
    \caption{Back-door criterion}
    \label{fig:back-door}
\end{figure}

\textbf{Back-door criterion.}
Fig.~\ref{fig:back-door} shows the back-door criterion, which targets for removing the intervention effects with \textit{do} operation. \textit{Do} operation denotes a surgery to cut off the connection from $C$ to $X$. From Fig.~\ref{fig:back-door}(a), the $P(Y|X)$ is associated with two paths, respectively as $X\xleftarrow{}C\xrightarrow{}Y$ and $X\xrightarrow{}Y$. Here $X\xleftarrow{}C\xrightarrow{}Y$ is a spurious path, which intervenes the estimation of causality between $X$ and $Y$, denoted as $P(Y|do(X))$ (\ieno, the Fig.~\ref{fig:back-door}(b)). Following the \citet{pearl2016causal}, we can denote the conditional probability between $X$ and $Y$ in Fig.~\ref{fig:back-door}(b) as $P_m(Y|X)$. Since $C$ has no parent-level variable, the $P_m(C)=P_m(C|X)$ in the Fig.~\ref{fig:back-door}(b) is equivalent to the $P(C)$ in the Fig.~\ref{fig:back-door}(a). Furthermore, we can get the $P_m(Y|X, C)$ in the Fig.~\ref{fig:back-door}(b) is equivalent to  $P(Y|X, C)$ in the Fig.~\ref{fig:back-door}(a) since the same graph architecture. 
Based on the above definition, the back-door criterion can be derived as:
\begin{align}
\centering
    P(Y&|do(X))=P_m(Y|X)\notag \\
    &=\sum_{c}P_m(Y|X, C=c)P_m(C=c|X) \notag \\
    &=\sum_{c}P_m(Y|X, C=c)P_m(C=c) \notag \\
    &=\sum_{c}P(Y|X, C=c)P(C=c)
    \label{eq:back-door}
\end{align}
Then we can get the formulation of the back-door criterion as:
\begin{equation}
    P(Y|do(X))=\sum_{c}P(Y|X, C=c)P(C=c).
\end{equation}
\begin{figure}[t]
    \centering
\includegraphics[width=0.6\linewidth]{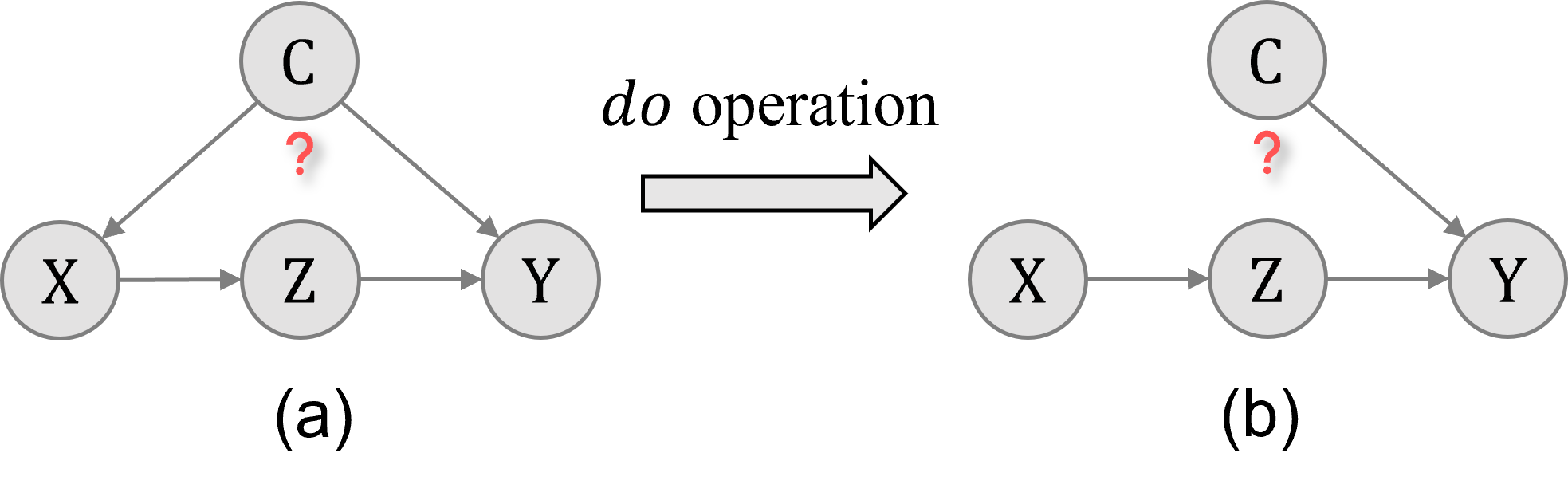}
    \caption{Front-door criterion}
    \label{fig:front-door}
\end{figure}
\textbf{Front-door criterion.}
From Eq.~\ref{eq:back-door}, we can draw a conclusion that estimating the causality between $X$ and $Y$ (\ieno, $P(Y|do(X)$) requires traversing the distribution of confounders $C$. However, confounders $C$ are in general diverse and not identifiable. To solve the above question, as shown in Fig.~\ref{fig:front-door}, the front-door criterion introduces an intermediate variable $Z$ and transfers the requirement of modeling the intervening effects of confounders $C$ on $X\xrightarrow{}Y$ to modeling the intervening effects of $X$ on $Z\xrightarrow{}Y$~\citep{pearl2016causal}.  Specifically, front-door criterion decompose the $P(Y|do(X))$ into two components, \ieno,  $P(Y|do(Z))$ and $P(Z|do(X))$, which can be represented as:
\begin{equation}
    P(Y|do(X))=\sum_{z}P(Z=z|do(X))P(Y|do(Z=z))
    \label{eq:13}
\end{equation}
As shown in Fig.~\ref{fig:front-door}(a), the variables $X$ and $Z$ do not have common causes, which reveals the path $X\xrightarrow{}Z$ is not intervened by other variables. Therefore, the causality between $X$ and $Z$ is equivalent to its correlation as:
\begin{equation}
     P(Z=z|do(X))=P(Z=z|X)
     \label{eq:14}
\end{equation}
From Fig.~\ref{fig:front-door}(a), the path $Z \xrightarrow{} Y$ is intervened by two variables, respectively as $C$, $X$, since the existing spurious path $Z\xleftarrow{}X \xleftarrow{}C \xrightarrow{}Y$. Then we can simply block this spurious path by cutting off the path $X\xrightarrow{}Z$ with back-door criterion~\citep{pearl2016causal}, which gets rid of identifying the confounders $C$. 
\begin{equation}
    P(Y|do(Z=z))=\sum_{\Tilde{x}\in X}P(Y|Z=z, \Tilde{x})P(\Tilde{x})
    \label{eq:15}
\end{equation}
Based on Eq.~\ref{eq:14} and~\ref{eq:15}, we can derive Eq.~\ref{eq:13} as: 
\begin{equation}
    P(Y|do(X))=\sum_{z}P(Z=z|X)\sum_{\Tilde{x}\in X}P(Y|Z=z, \Tilde{x})P(\Tilde{x})
    \label{eq:16}
\end{equation}
Then we can obtain the formulation of the front-door criterion as Eq.~\ref{eq:16}.

\subsection{\tcr{Using $h(X)$ as $Z$} satisfies front-door criterion}
\label{sec:proof}
In this section, we give the proof that using $h(x)$ as $Z$ satisfies the front-door criterion. As depicted in \citet{pearl2016causal}, the definition of the front-door criterion is as following:

\textbf{Definition:} \textit{if $Z$ satisfies the front-door criterion relative to an ordered pair of variables $(X, Y)$, it must obey the following principles: 1) Z intercepts all directed paths from $X$ to $Y$. 2) There is no unblocked back-door path from $X$ to $Z$. 3) All back-door paths from $Z$ to $Y$ are blocked by $X$.} 

We prove that ours satisfy the above principles as:
\begin{itemize}
    \item As shown in Fig.~\ref{fig:SCM_for_Training}, the direct path from $X$ to $Y$ is built from the model $\varphi$, that is composed of two sub-models, $h$ and $f$. We can represent this path as $Z=g(X), Y=f(Z)$. We can observe that $Z$ intercepts all directed paths from $X$ to $Y$, which satisfies the \textbf{\textit{Principle 1)}}. 
    \item The second principle requires there is no unblocked back-door path from $X$ to $Z$, (\textit{i.e.,} $P(Z|do(X))=P(Z|X)$), which is a vital factor in ensuring that the causal effects can be estimated. We interpret that ours satisfy this principle from two perspectives. 1) As shown in Fig.~\ref{fig:SCM_for_Training}, the effect of confounders C to Z is transmitted
by X through a sub-model $f$ as $C\xrightarrow{}X \xrightarrow[]{f}Z$ instead of $X\xleftarrow[]{}C\xrightarrow[]{}Z$, which indicates that there is no
back-door from X to Z and thus no common confounders for X and
Z. 2) We prove this from contradiction. From the theory of
the back-door criterion, if there are common confounders C for X and Z,
then $P(Z|do(X))! = P(Z|X)$. In fact, when $Z = h(X)$
is obtained from a deterministic mapping of X at the inference stage, we always have
$P(Z|X, C) = P(Z|X)$ for any confounder C. Then we
can derive that $P(Z|do(X)) = \sum_{C=c} P(Z|X, c)P(c) =
\sum_{C=c} P(Z|X)P(c) = P(Z|X)$. This means there are no common confounders for X and Z (\textit{i.e.,} no unblocked back-door path for X and Z), which satisfies the \textbf{\textit{Principal 2)}}.
\item There are two paths from $Z$ to $Y$, \textit{i.e.,} $Z\xrightarrow[]{}Y$ and $Z\xleftarrow[]{}X\xleftarrow[]{}C\xrightarrow[]{}Y$. The second path is called the back-door path between $Z$ and $Y$. When we condition on $X$, the back-door path will be blocked. Therefore, all back-door paths from $Z$ to $Y$ are blocked by $X$, which satisfies the \textbf{\textit{Principal 3)}}. 
\end{itemize}
Consequently, exploiting $Z=h(X)$ is consistent with the front-door criterion.  

Another essential factor for estimating the causal effects by the front-door criterion is $P(X, Z)>0$. That means if $P(X, Z)$=0, the $P(Y|do(X))$ in Eq.~\ref{equ:front-door} is equivalent 0 and cannot be estimated. 
In our \textbf{CICF}, $Z$ is deterministic related to $X$
with $Z=f(X)$. Based on the probability theory $P(X, Z) = P(X)P(Z|X)$. Since $Z=f(X)$, the $P(Z|X)=P(f(X)|X)=1$ is always holds when $f$ is fixed at inference stage. Therefore, $P(X, Z)=P(X)P(Z|X)=P(X)>0$ holds in our method. 

\subsection{Different choices of $Z$}
\label{sec:choicez}
\begin{figure}[htp]
\centering
    \includegraphics[width=0.5\linewidth]{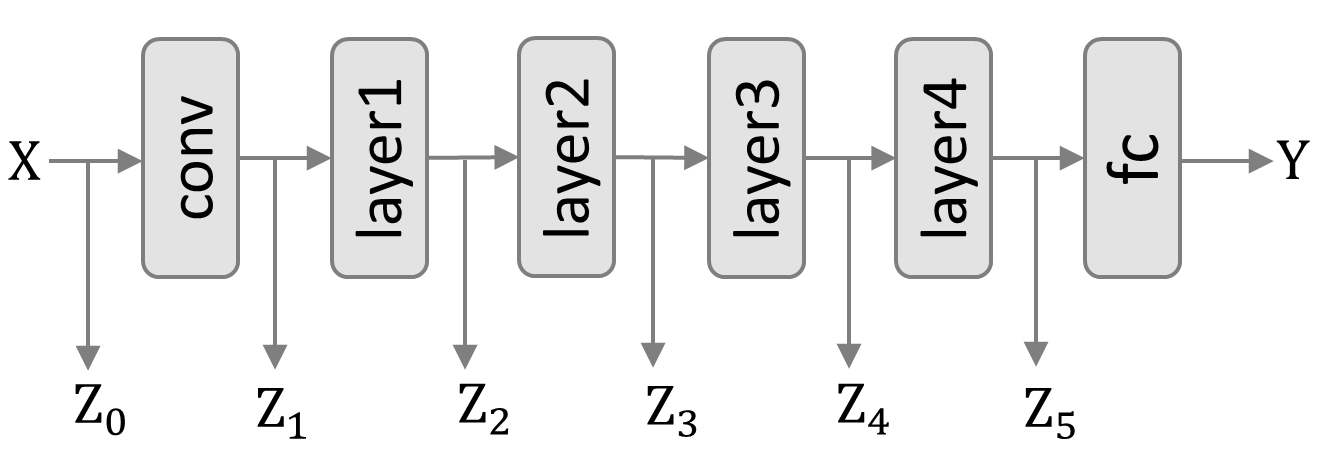}
    \caption{Different choices for the intermediate output $Z$.}
    \label{fig: ablation_layer}
\end{figure}

As shown in Fig.~\ref{fig:SCM_for_Training} (a), the model $\varphi$ is separated into successive $h$ and $f$, and $Z=h(X)$ is the intermediate output of $\varphi$.
To explore the effects of different choices of $Z$, \ieno, and different separations of $\varphi$, we conduct ablation experiments on PACS. Fig.~\ref{fig: ablation_layer} shows the different choices of $Z$ based on ResNet and the corresponding results are shown in Fig. \ref{fig:ablation}(c). 
It is observed that the shallower $Z$ is obtained, the better accuracy the results achieved. We reckon the reason is that selecting $Z$ from shallower layers will result in a larger model for $f$, which will have more capability to learn the conditional probability $P(Y|do(Z))$. In this paper, we set $Z=Z_0$ to make $f$ have more parameters to learn causal features.

\subsection{More Implementation Details}
\label{sec:more_implementation}
For PACS and Office-Home, we take ResNet18~\citep{he2016deepResNet} pretrained on ImageNet~\citep{deng2009imagenet} as backbone, following~\citet{zhou2021domainMixstyle,carlucci2019domain,li2018learningMLDG}. For VLCS, we take AlexNet~\citep{krizhevsky2012imagenetAlexNet} pretrained on ImageNet~\citep{deng2009imagenet} as our backbone, which is the same as ~\citet{dou2019domainMASF,li2019episodic,matsuura2020domainMMLD}. For Digits-DG, we adopt the model architecture used in previous works \citep{zhou2020learningL2A-OT,carlucci2019domain,zhou2021domainMixstyle},
which is composed of four convolution layers with inserted ReLU and max-pooling layers. 
\begin{figure*}[t]
    \centering
    \includegraphics[width=0.95\linewidth]{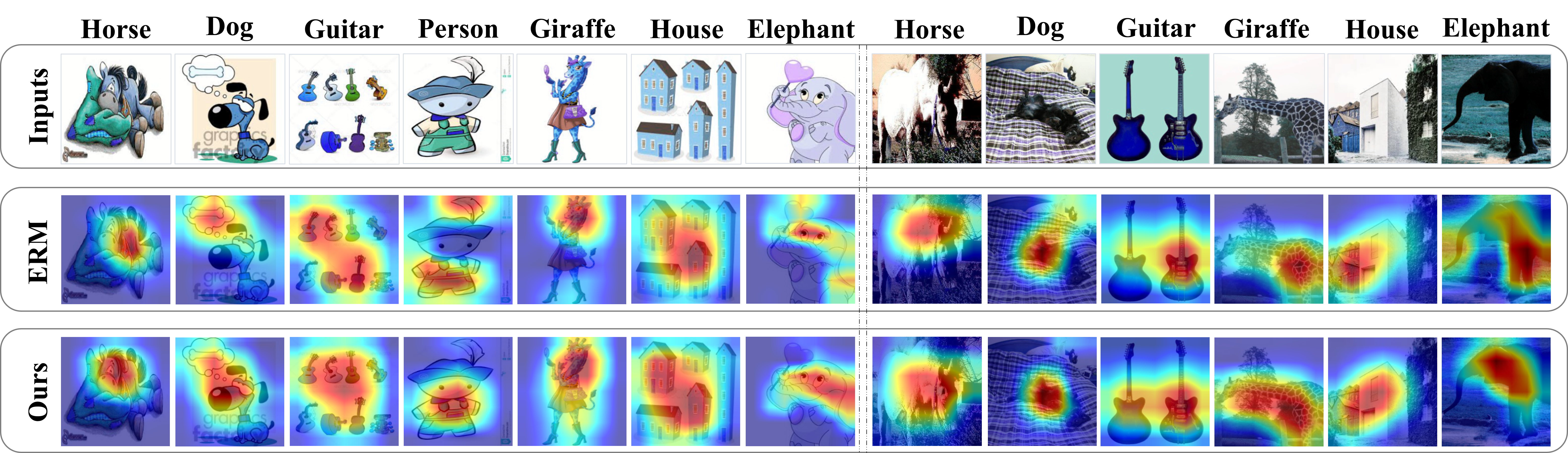}
    \caption{Visualization of the Grad-Cams w.r.t. classification task for domain generalization. The first row of left/right panels shows the images from cartoon/photo domains in PACS. The second and third rows show the Grad-Cams of ERM and our \ours.}
    \label{fig:gradcam_supplementary}
\end{figure*}
\begin{figure}[t]
    \centering
    \includegraphics[width=0.95\linewidth]{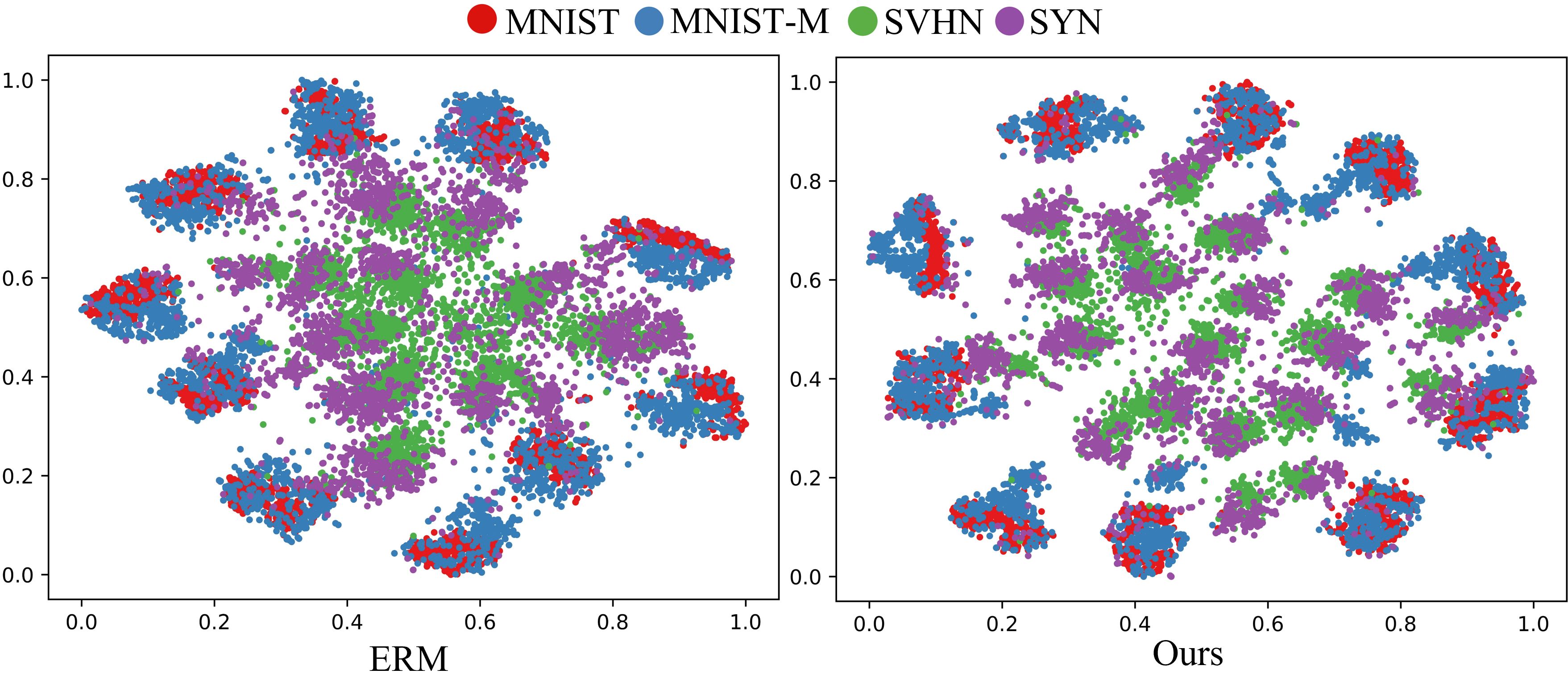}
    \caption{t-SNE visualization of features learned by ERM (left) and our \ours~(right).}
    \label{fig:tsne}
\end{figure}
We set the mini-batch size $M$ for computing the global-scope gradient as 256,
and the mini-batch size $M^l$ for computing $L_\ours$ as 84. 
We set the batch size to 84 and train the model using SGD optimizer for 60 epochs. For PACS, the learning rate $\alpha$ and $\beta$ are set to 0.05 and 0.01 respectively. For VLCS,  the learning rate $\alpha$ and $\beta$ are 0.005 and 0.001 respectively. As for Office-Home, we set $\alpha$, $\beta$ as 0.001 and 0.001 respectively. For Digits-DG, we set $\alpha$, $\beta$ as 0.5 and 0.1 respectively. All reported results are averaged among six runs with different seeds.

The GPU memory cost of our \textit{CICF} is 7.151 GiB and the clustering for training data takes 164 seconds for PACS, which introduces only $4.8\%$ extra time to the whole training process (\ieno, 3432 seconds) while bringing the significant gain of 2.4\%.

\begin{table}[htp]
\centering
\caption{Classification accuracy ($\%$) of different DG methods on Office-Home with ResNet-18 as the backbone.}
\setlength{\tabcolsep}{2pt}{
\begin{tabular}{c|ccccc}
\hline
\multirow{2}{*}{Method} & \multicolumn{5}{c}{Office-Home}                 \\ \cline{2-6} 
                        & Art  & Clipart & Product & Realworld & Avg.  \\ \cline{1-6} 
JiGen                   & 53.0 & 47.5    & 71.5    & 72.8       & 61.2      \\
MMD-AAE                & 56.5 & 47.3    & 72.1    & 74.8       & 62.7      \\
MLDG & 57.8 &50.3 &70.6 &73.0 & 63.0 \\
CrossGrad               & 58.4 & 49.4    & 73.9    & 75.8       & 64.4       \\
CCSA                  & \textbf{59.9} & 49.9    & 74.1    & 75.7       & 64.9    \\
MixStyle                & 58.7 & 53.4    & 74.2    & \textbf{75.9}       & 65.5       \\ \hline
ERM               & 58.1 & 48.7    & 74.0    & 75.6       & 64.2  \\
ERM+MAML           & 56.8 & 52.5    & 74.0    & 74.7   & 64.5   \\
\rowcolor[gray]{0.9}
ERM+\ours          & 57.1 & 52.0    & 74.1    & 75.6       & 64.7     \\ \cline{1-6} 
ERM$^*$               & 59.6     & 53.0         & \textbf{74.3}        &  75.4          &  65.6      \\
ERM$^*$+MAML               & 56.2     & 56.1         & 72.6       &  73.2          &  64.5     \\
\rowcolor[gray]{0.9}
ERM$^*$+\ours           &  59.3    & \textbf{56.2}       & 74.2        &  75.1          & \textbf{66.2}       \\ \hline
\end{tabular}}
\label{tab: office-home}
\end{table}

\begin{table}[t]
\centering
\caption{Classification accuracy ($\%$) of different DG methods on VLCS with AlexNet as the backbone.}
\setlength{\tabcolsep}{2pt}{
\begin{tabular}{c|ccccc}
\hline
\multirow{2}{*}{Methods} & \multicolumn{5}{c}{VLCS}                 \\ \cline{2-6} 
                         & Caltech & Labelme & Pascal & Sun  & Avg. \\ \hline
MLDG
            & 97.9 & 59.5 &66.4 & 64.8 & 72.2 \\
Epi-FCR                 & 94.1    & 64.3    & 67.1   & 65.9 & 72.9 \\
JiGen                  & 96.93   & 60.9    & 70.6   & 64.3 & 73.2 \\
MMLD                     & 96.6    & 58.7    & \textbf{72.1}   & 66.8 & 73.5 \\
MASF                     & 94.8    & \textbf{64.9}    & 69.1   & 67.6 & 74.1 \\\hline
ERM                      &  96.3       &  59.7       & 70.6       &  64.5    &  72.8    \\
ERM+MAML                 &  97.8       &  58.0       & 67.1       &  64.1    &  71.8    \\
\rowcolor[gray]{0.9}
ERM+\ours               &  97.8       &  60.1       &  69.7    &    67.3  &   73.7   \\\hline
ERM$^*$                    & 96.4        &  60.7       &  68.6      &  66.2    &   73.0   \\
ERM$^*$+MAML                & 98.1        &  58.2       &  69.6      & 64.5     &  72.6    \\
\rowcolor[gray]{0.9}
ERM$^*$+\ours               &  \textbf{98.1}       &   62.4      &   69.3     &  \textbf{69.1}    &  \textbf{74.7}    \\ \hline
\end{tabular}}
\label{tab: vlcs}
\end{table}

\subsection{Feature Visualization}
\label{sec:more_feature_vis}
We visualize more Grad-CAM~\citep{selvaraju2017gradGradcam} of ERM and ERM+\ours~in Fig.~\ref{fig:gradcam_supplementary}. We can observe that our \ours~ focus more on foreground regions (\ieno, the casual features), while ERM easily focuses on the misleading regions (\egno, the bone in the dog of the cartoon, background) when capturing causal features. 
As shown in Fig.~\ref{fig:tsne}, we visualize the learned feature on Digits-DG \textbf{by t-SNE}~\citep{saito2019strongTSNE}. We find that the distribution of features extracted from ERM+\ours~is more compact across samples with the same category, compared to ERM. This validates the effectiveness of our algorithm for causal feature learning.

\end{document}